\documentclass{article}

\PassOptionsToPackage{numbers}{natbib}  % ← 加这一行

\usepackage{amsmath}
\usepackage[preprint]{neurips_2025}

\usepackage[utf8]{inputenc} % allow utf-8 input
\usepackage[T1]{fontenc}    % use 8-bit T1 fonts
\usepackage{hyperref}       % hyperlinks
\usepackage{url}            % simple URL typesetting
\usepackage{booktabs}       % professional-quality tables
\usepackage{amsfonts}       % blackboard math symbols
\usepackage{nicefrac}       % compact symbols for 1/2, etc.
\usepackage{microtype}      % microtypography
\usepackage{xcolor}         % colors
\usepackage{tabularx}
\usepackage{graphicx}
\usepackage{enumitem}
\usepackage{graphicx}
\usepackage{enumitem}
\usepackage{subcaption}
\usepackage{wrapfig}
\usepackage{listings}
\usepackage{algorithm}
\usepackage{algpseudocode}
\usepackage[most]{tcolorbox}
\usepackage{xcolor}
\usepackage{listings}
\usepackage{booktabs}
\usepackage{longtable}
\usepackage{array}
\usepackage[utf8]{inputenc}

\newcolumntype{P}[1]{>{\raggedright\arraybackslash}p{#1}}
\newcommand{\scoreitem}[2]{\textbf{#1}: #2\par\vspace{0.25em}}

\definecolor{promptbg}{RGB}{248,248,248}
\definecolor{promptframe}{RGB}{210,210,210}
\definecolor{prompttitle}{RGB}{60,60,60}

\lstdefinestyle{promptstyle}{
  basicstyle=\footnotesize\ttfamily,
  breaklines=true,
  breakatwhitespace=false,
  columns=fullflexible,
  keepspaces=true,
  showstringspaces=false,
  tabsize=2,
  xleftmargin=0.5em,
  xrightmargin=0.5em
}

\newtcblisting{promptbox}[2][]{
  enhanced,
  breakable,
  listing only,
  listing style=promptstyle,
  colback=promptbg,
  colframe=promptframe,
  coltitle=white,
  fonttitle=\bfseries,
  title=#2,
  boxed title style={
    colback=prompttitle,
    colframe=prompttitle,
    arc=1mm
  },
  attach boxed title to top left={xshift=2mm,yshift=-2mm},
  top=3mm,
  bottom=2mm,
  left=1mm,
  right=1mm,
  arc=1mm,
  boxrule=0.4pt,
  #1
}

\title{CuraWeb: Joint Optimization of Quality, Redundancy, and Diversity for Web-Scale Pretraining Data}
\author{
\textbf{Peiguang Li\textsuperscript{1}}\thanks{Equal contribution.}
\quad
\textbf{Yongwei Zhou\textsuperscript{1}}\footnotemark[1]
\quad
\textbf{Juncheng Diao\textsuperscript{1}}\footnotemark[1]
\quad
\textbf{Yuchun Fan\textsuperscript{1}}\footnotemark[1]
\quad
\textbf{Jian Yang\textsuperscript{1}}
\\
\textbf{Jianxiao Yang\textsuperscript{1}}
\quad
\textbf{Zhongda Su\textsuperscript{1}}
\quad
\textbf{Shuguang Jiao\textsuperscript{1}}
\quad
\textbf{Xiao Wei\textsuperscript{1}}
\quad
\textbf{Zhiye Zou\textsuperscript{1}}
\\
\textbf{Gan Dong\textsuperscript{1}}
\quad
\textbf{Zhizhao Zeng\textsuperscript{1}}
\quad
\textbf{Rongxiang Weng\textsuperscript{1}}\thanks{Corresponding authors.}
\quad
\textbf{Jingang Wang\textsuperscript{1}}\footnotemark[2]
\quad
\textbf{Xunliang Cai\textsuperscript{1}}
\\[0.5em]
\textsuperscript{1}Meituan
\\[0.3em]
{\small \texttt{lipeiguang17@mails.ucas.ac.cn},\quad \texttt{ywzhouphd2018@gmail.com},\quad 
\texttt{diaojuncheng24@mails.ucas.ac.cn}}
\\[0.1em]
{\small \texttt{yuchunfan\_neu@outlook.com},\quad \texttt{wengrongxiang@meituan.com},\quad \texttt{wangjingang02@meituan.com}}
}

\begin{document}

\maketitle
\begin{abstract}
Open-web corpora curated via highly selective filters, such as FineWeb-Edu and DCLM, constitute the core of LLM pretraining data and have significantly advanced LLM performance. However, these pipelines typically rely on singular optimization objectives, which inevitably narrows distributional diversity and marginalizes long-tail knowledge, thereby restricting data coverage and underutilizing the vast potential of the open web.
To address this limitation, we propose a novel curation paradigm that shifts from linear pruning to the joint optimization of quality, redundancy, and diversity.
This framework synergizes dual-track cleaning (rule-based and model-driven) with hybrid deduplication (n-gram and semantic), while employing a multi-objective sampler to balance informational quality with distributional breadth. Applying this framework to Common Crawl, we construct \textsc{CuraWeb}, a 2T-token English corpus. Unlike existing resources, 
\textsc{CuraWeb} establishes an industrial-grade standard for data curation 
by recovering a more holistic data distribution with enhanced diversity and 
minimal redundancy, achieving broader coverage of long-tail knowledge across 
diverse domains.
Experimental evaluations at the 3B scale demonstrate that \textsc{CuraWeb} significantly outperforms state-of-the-art baselines, yielding an average performance gain of 1.8\% across a wide range of benchmarks, particularly in knowledge-intensive and reasoning tasks.
\end{abstract}

\section{Introduction}                           
In the current paradigm of Large Language Models (LLMs) pre-training, the quality and diversity of web-scale corpora, such as Common Crawl, are recognized as the core variables determining the upper bound of a model's emergent capabilities and reasoning performance \cite{xi2025samplemix,achiam2023gpt,liu2024deepseek,yang2025qwen3,team2025longcat,grattafiori2024llama,adler2024nemotron}. 
However, prevailing data curation frameworks, such as FineWeb-Edu \cite{penedo2024fineweb} and DCLM \cite{li2024datacomp}, traditionally rely on a singular optimization approach, deploying monolithic quality classifiers to filter data. 
Our research indicates that while this paradigm effectively enhances general stylistic quality, it inevitably narrows distributional diversity. 
Furthermore, existing pipelines treat curation as isolated, sequential stages---cleaning, deduplication, and sampling---each driven by the same uni-dimensional quality signal. 
This structural isolation causes the same distributional bias to compound across the entire pipeline, leaving quality-diversity trade-offs systematically unresolved. 
Together, these limitations render such frameworks inadequate for the rigorous requirements of industrial-scale pre-training.

To address these limitations, we propose a unified data curation framework that shifts the paradigm from disjointed, stage-wise filtering---namely, isolated cleaning, deduplication, and sampling---toward multi-dimensional signal-driven global governance. 
Specifically, we train specialized models to annotate all pre-cleaned data with fine-grained signals, including multi-aspect quality scores and domain tags. 
Crucially, these multi-dimensional attributes serve as core dependencies across the entire pipeline, enabling consistent funnel monitoring and strategy optimization. 
This holistic approach ensures a well-calibrated balance between data quality, domain coverage, and token diversity. 

To operationalize this holistic approach, our framework introduces three core technical innovations:
First, we refine traditional heuristic-based filtering \cite{rae2021scaling,penedo2023refinedweb,wenzek2020ccnet} by replacing uniform constraints with domain-specific thresholds, thereby preventing the inadvertent removal of high-value samples in rigorous fields like mathematics and code. 
To complement these calibrated heuristics, we further incorporate writing quality and content value signals derived from our data understanding model to perform granular preliminary filtering. 
Together, these domain-aware heuristics and multi-dimensional quality signals enable the retention of significantly more high-value data than conventional baselines, all while preserving training convergence efficiency.

Second, we introduce a hybrid deduplication pipeline to address deep semantic redundancy that evades standard $n$-gram filters. 
We identify two critical types of hidden redundancy: \textit{Templatized Clusters} (documents sharing a structural skeleton but differing in localized parameters) and \textit{Redundant Crawling} falling just below the Locality-Sensitive Hashing (LSH) threshold. 
To handle these boundary cases, we propose a soft semantic deduplication strategy based on a weighted voting mechanism. 
By accumulating penalty scores across multiple similarity edges, this approach classifies a document as redundant only when its aggregated score exceeds a predefined threshold. 
This strategy avoids the aggressive over-filtering of hard-threshold approaches \cite{huo2025dots}, reducing the false positive rate from $37.5\%$ to $28.03\%$ in the highest similarity interval.

Third, we introduce an automated sampling framework that jointly optimizes quality and diversity. 
Documents are evaluated along two complementary dimensions: writing quality (e.g., coherence and logic), which is quantified via a harmonic mean to heavily penalize any severe single-dimension flaw; and content value, scored via linear summation to reward specialized expertise and high information density. 
This combined quality metric is further integrated with a diversity score to ensure broad domain representation. 
To prioritize highly informative content, we adopt Power Sampling, which amplifies the selection probability of top-tier documents.

Experimental results on a 1T-token scale demonstrate that \textsc{CuraWeb} significantly outperforms strong baselines like FineWeb-Edu, Dolma3, and DCLM, particularly in knowledge-intensive and reasoning tasks \cite{cobbe2021training,amini2019mathqa,hendrycks2021measuring,wang2024mmlu,lai2017race,zellers2019hellaswag,bisk2020piqa,welbl2017crowdsourcing,sakaguchi2021winogrande,mihaylov2018can,olmo2025olmo3}. 
The robust performance gains across diverse benchmarks validate that our full-stack collaborative optimization effectively maximizes the token efficiency of the training corpus. 
This provides strong empirical evidence that meticulous data refinement and distribution reshaping are critical paths for advancing the intelligence of LLMs.

\section{Related Work}      

\paragraph{Coarse-Grained Data Filtering}
A significant portion of raw Common Crawl data consists of low-quality and duplicate content, requiring rule-based filtering and deduplication. 
Rule-based filtering has steadily evolved from surface-level text statistical features to deep structured extraction. 
Early frameworks such as CCNet introduced language identification and KenLM perplexity scoring \cite{wenzek2020ccnet}, while C4 applied line-level heuristics to remove residual HTML markup and boilerplate noise \cite{dodge2021documenting}. 
More recently, RefinedWeb and FineWeb2 leverage complete HTML structures to achieve more precise text extraction \cite{penedo2023refinedweb,penedo2025fineweb}. 
For deduplication, $n$-gram fingerprint matching via MinHash-LSH remains dominant \cite{broder1997resemblance,indyk1998approximate}, with recent studies extending to embedding-based semantic deduplication through vector databases to capture deeper content redundancy \cite{abbas2023semdedup}. 
However, these approaches suffer from two fundamental limitations: first, rule-based filters apply uniform thresholds across all domains, inadvertently discarding high-value content in specialized fields such as mathematics and science; second, existing deduplication methods rely on fixed hard thresholds for binary filtering, incurring the risk of over-filtering domain-specific documents that naturally exhibit dense terminological overlap \cite{lee2022deduplicating,abbas2023semdedup,shen2023slimpajama}. 
\textsc{CuraWeb} addresses both limitations by introducing domain-aware heuristic thresholds and a soft semantic deduplication strategy based on a weighted voting mechanism, thereby preserving high-value long-tail content while maintaining robust deduplication effectiveness.

\paragraph{Fine-Grained Content Optimization}
Web corpora that have undergone coarse-grained filtering still suffer from excessive low-value content and severe domain distribution imbalances. 
Consequently, content quality assessment has shifted from rule-driven paradigms to sophisticated model-driven approaches. 
Earlier protocols, such as GPT-3 and LLaMA, trained binary classifiers on high-quality reference corpora to distinguish premium documents \cite{brown2020language,touvron2023llama}, whereas RefinedWeb constructed a multi-dimensional scoring system integrating URL authority, content length, and structural features \cite{penedo2023refinedweb}. 
FineWeb-Edu and DCLM pushed this trajectory further by employing monolithic quality classifiers trained on highly curated reference sets, achieving strong downstream benchmark performance, albeit at the expense of narrowing distributional diversity \cite{penedo2024fineweb,li2024datacomp}. 
To regulate domain distributions, techniques like doremi deploy weighted sampling to amplify the probability of selecting high-quality or low-resource domain documents \cite{xie2023doremi}, while the recently proposed WebOrganizer framework introduces automated topic and format classification for precise domain mixing \cite{wettig2025organize}. 
However, these methods traditionally treat quality and diversity as disjoint objectives optimized in isolation, leaving the inherent trade-off between them systematically unresolved. 
\textsc{CuraWeb} reconciles this tension by unifying multi-aspect quality and diversity signals into a shared multi-dimensional framework, enabling globally consistent joint optimization across the entire data curation pipeline.

\section{The \textsc{CuraWeb} Curation Framework}

\subsection{Overview}

This section details the design and composition of the CuraWeb dataset. 
It outlines the end-to-end pipeline, starting with raw text extraction and parsing, followed by the construction of a comprehensive pre-training data understanding system. 
Building on this foundation, we elaborate on the subsequent stages of data refinement, including cleaning, deduplication, selection, and sampling. 
This methodology ensures a rigorous balance between data quality, diversity, and domain coverage.

\begin{figure*}[t]
  \centering
  \includegraphics[width=0.95\textwidth]{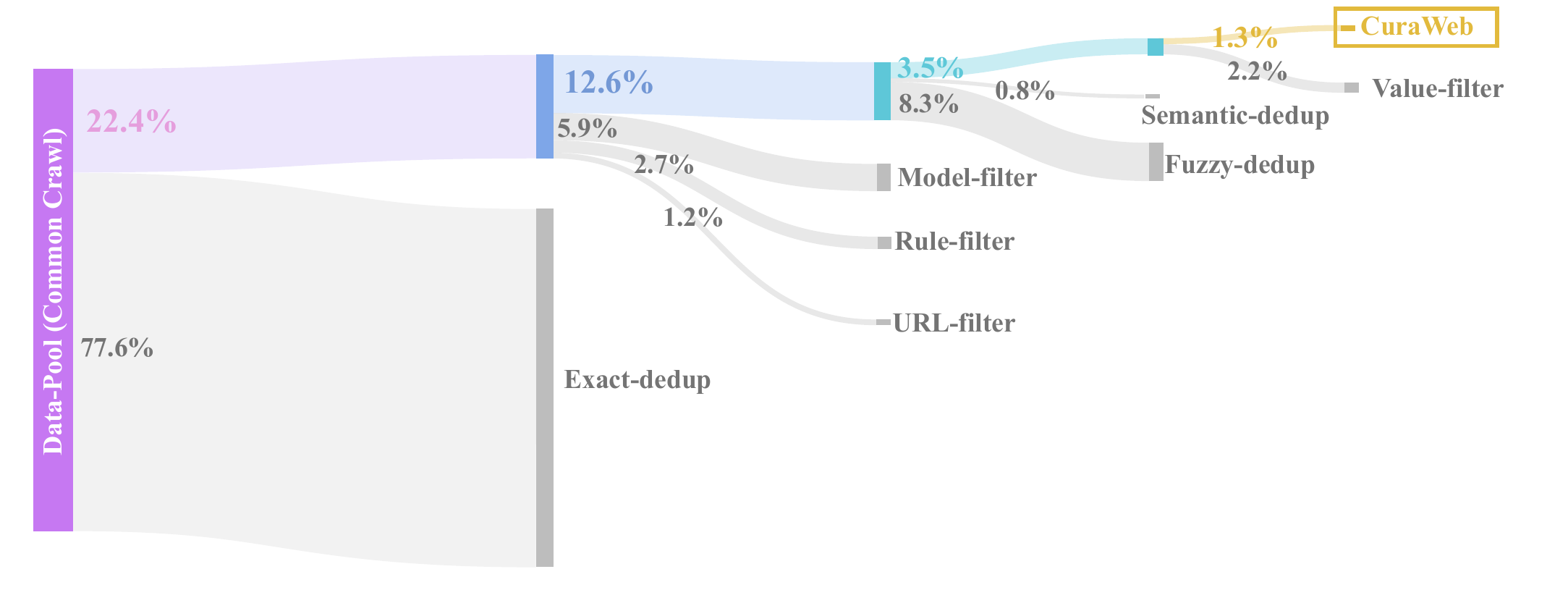}
  \caption{
 Construction pipeline of \textsc{CuraWeb}.
The funnel shows how raw common crawls are progressively refined through exact-deduplication, filtering, deduplication, and data selection.
Percentages denote the relative document flow or removal ratio at each stage.
  }
  \label{fig:funnel_pipeline}
\end{figure*}

\subsection{Text Extraction}
To ensure comparability with established baselines, we follow the preprocessing protocols of DCLM \cite{DCLM-2024} and FineWeb \cite{penedo2024fineweb}. 
CuraWeb is constructed from twelve years of public Common Crawl snapshots spanning 2013 to 2024. Main-body text is extracted from raw WARC files using Resiliparse \cite{bevendorff2018elastic}, removing irrelevant headers, footers, and advertisement noise. 
A pre-trained FastText language classifier\cite{joulin2017bag} then retains only English documents with a confidence score exceeding 0.5. 
For exact deduplication, we apply global SHA-256 document hashing, compressing 184.9B raw documents to 41.4B unique documents (retention rate: 22.4\%). Finally, an extended URL blacklist—integrating public open-source lists and self-built filtering rules—filters spam and adult content domains, yielding a final collection of 39.1B documents (retention rate: 94.5\%). 
The complete data funnel statistics are reported in Figure\ref{fig:funnel_pipeline}.

\subsection{Data Understanding System}
\label{sec:Data_Understanding}

Traditional pre-training data optimization is commonly conducted in a black-box manner, lacking explicit supervision over the entire data processing pipeline\cite{longpre2024pretrainer,DCLM-2024,penedo2024fineweb}. In large-scale scenarios, this can easily induce uncontrollable data distribution shifts, making systematic biases difficult to detect and rectify. 
To address this, we perform multi-signal annotation on extracted text prior to cleaning, constructing a feature system that serves as a unified observation benchmark throughout the pipeline.
This system plays two key roles: it enables quantitative analysis of filtering behaviors toward specific domains (e.g., long-tail STEM content), facilitating the detection of systematic biases; and it provides feature guidance for subsequent sampling algorithms to dynamically adjust weights based on domain and quality distributions.

To comprehensively characterize web text features for full-pipeline supervision, we construct a multi-dimensional signal system consisting of two core components:

\begin{wrapfigure}{r}{0.46\textwidth}
  \vspace{0pt}
  \centering
  \includegraphics[width=0.44\textwidth]{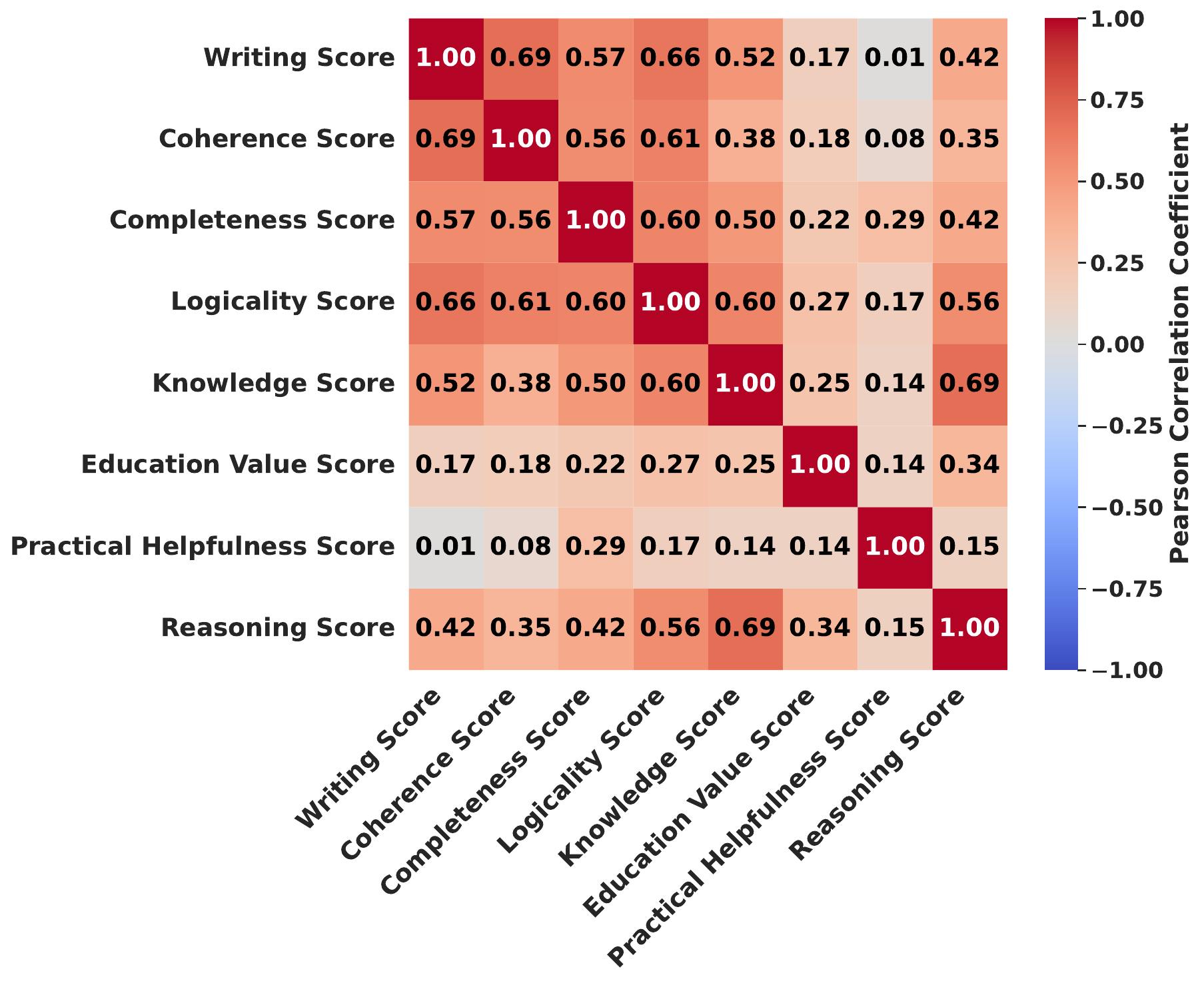}
  \caption{
  Pearson Correlation heatmap displaying across the derived quality metrics. Note that the \textit{safety\_score} metric (binary) is excluded from this heatmap due to its scale variance from the other 4-point scale metrics.
  }
  \vspace{-1.2em}
  \label{fig:nmi_label_correlation}
\end{wrapfigure}

\textbf{Quality Metrics}:
To capture multi-faceted text characteristics beyond a single monolithic quality score—which frequently masks localized document defects—we define nine fine-grained evaluation dimensions. These metrics encompass \textit{writing\_score}, \textit{coherence\_score}, \textit{completeness\_score}, \textit{logicality\_score}, \textit{safety\_score}, \textit{knowledge\_score}, \textit{education\_score}, \textit{helpfulness\_score}, and \textit{reasoning\_score}. 
To verify the statistical orthogonality of these dimensions, we compute the Pearson across the annotated corpus. As illustrated in the low-collinearity heatmap in Fig.~\ref{fig:nmi_label_correlation}, these metrics exhibit high feature independence, ensuring distinct information representation.

\textbf{Domain Taxonomies}:
Domain signals map documents to a two-level thematic taxonomy adapted from {Google AdSense}'s classification hierarchy.
Specifically, we sample over one million web documents and employ GPT-4o to perform automatic categorization, followed by merging of semantically similar categories, resulting in a final taxonomy of 26 coarse-grained themes and 105 fine-grained categories covering a broad range of topics across the English Web.

To minimize the computational overhead of annotating billions of documents while maintaining high annotation precision, we develop a lightweight multi-task learning model with a 300M-parameter Transformer backbone. Two independent linear heads are attached to the representation layer to perform quality assessment and domain classification in a single forward pass.
To boost its performance, we adopt a unified knowledge distillation pipeline for all tasks.
We first conduct stratified sampling to select 500K documents from the raw corpus, and employ GPT-4o \cite{hurst2024gpt} as the teacher model to generate high-quality supervision labels for the two annotation tasks. These supervisory signals are then distilled to train our lightweight student model. To alleviate gradient imbalance among multiple classification subtasks during joint optimization, we introduce an uncertainty weighting strategy \cite{kendall2018multi} that dynamically adjusts the loss weight of each task, facilitating stable and balanced model convergence.

\subsection{Data Filtering}
\label{sec:data_filtering}

We adopt a two-stage filtering pipeline combining heuristic rule-based filtering and model-based quality filtering. 
Different from conventional filtering strategies that simply remove noisy content, our pipeline aims to eliminate low-quality documents while preserving high-value long-tail content as a core objective.

\subsubsection{Rule Filter}
\label{sec:rule_filter}
Although heuristic rule pipelines such as Gopher\cite{rae2021scaling} have been widely adopted, their thresholds are calibrated on general web text, leading to systematic over-filtering of domain-specific content. 
For instance, the alphabetic character ratio rule—designed to filter garbled pages—inadvertently discards high-value STEM documents containing mathematical formulas and LaTeX expressions, resulting in systematic loss of scarce long-tail technical knowledge. 
Similar issues arise across multiple rules, motivating a systematic recalibration of the rule set.

To address these drawbacks, we evaluate each rule quantitatively by its filtering precision on sampled data, assessed through model-aided annotation and manual verification. As presented in Table~\ref{tab:rule_optimization}, we systematically optimize the conventional heuristic rule set in two steps. First, we remove five rules with low filtering precision or high risk of over-filtering domain-specific content. Second, we recalibrate the thresholds of three rules prone to over-filtering STEM content, as detailed in Table~\ref{tab:rule_optimization}. Furthermore, to avoid the accidental elimination of high-value professional content, we establish a priority bypass mechanism: documents assigned to Mathematics, STEM, and Code categories—as identified by the annotation model in Section~\ref{sec:Data_Understanding}—skip the heuristic filtering stage and are directly forwarded to the subsequent model-based filtering module.

\begin{table}[htbp]
\centering
\small
\caption{Summary of Heuristic Rule Modifications}
\label{tab:rule_optimization}
\resizebox{\columnwidth}{!}{
\begin{tabular}{llll}
\toprule
\textbf{Rule Name} & \textbf{Original} & \textbf{Optimized} & \textbf{Motivation} \\
\midrule
\texttt{median\_word\_length}   & $\le 3$   & \textit{Removed} & Reduces misjudgment of short-word technical texts \\
\texttt{required\_word\_count}  & $\le 2$   & \textit{Removed} & Eliminates redundant low-precision constraint \\
\texttt{transformed\_text\_len} & $\ge 10000$ & \textit{Removed} & Prevents purging of long technical specifications \\
\texttt{fraction\_bullet\_points} & $\ge 0.9$ & \textit{Removed} & Preserves well-structured technical outlines \\
\texttt{fraction\_ellipses}     & $\ge 0.3$ & \textit{Removed} & Removes low-precision constraint \\
\texttt{word\_count}            & $\le 50$  & $\mathbf{\le 25}$ & Relaxes word limit to retain concise knowledge snippets \\
\texttt{symbol\_to\_word\_ratio} & $> 0.1$  & $\mathbf{> 0.38}$ & Tolerates symbol-dense text to protect formulaic content \\
\texttt{alphabetic\_char\_ratio} & $< 0.8$  & $\mathbf{< 0.26}$ & Relaxes alphabetic constraint to prevent STEM data loss \\
\bottomrule
\end{tabular}
}
\end{table}

Our rule-based filtering achieves an overall precision of 90\%, and the recall rate for high-quality documents stays between 90\% and 95\% on our sampled evaluation set. 
This stage reduces the dataset from 39.1B to 34.2B documents, which means 87.3\% of the documents are kept. 
In contrast, if we use the traditional rule set with its original uniform thresholds, it only keeps 12.7B documents, leading to a much lower retention rate of only 32.6\%. 
This large difference clearly proves that traditional uniform rules filter out too much useful web data.

\subsubsection{Model Filter}
While rule-based filtering efficiently removes structurally malformed content, 
it operates on surface-level heuristics and thus faces two inherent limitations. 
First, relaxing rule thresholds to preserve data diversity inevitably allows 
more low-quality documents to pass through. 
Second, semantically deficient content---such as incoherent narratives, 
logically inconsistent arguments, or incomplete passages---cannot be reliably 
identified by rule-based signals alone.

To address these limitations, we adopt a cascaded filtering paradigm: we first apply rule-based filtering, and then perform model-based filtering for fine-grained quality assessment in the subsequent stage.
Here we utilize the multi-dimensional quality scores produced by the annotation model presented in Section~\ref{sec:Data_Understanding}.
Specifically, we assess each document along four dimensions most indicative of surface-level 
text quality: \textit{writing\_score}, \textit{coherence\_score}, 
\textit{completeness\_score}, and \textit{safety\_score}. 
We adopt a precision-oriented conservative strategy: any document scoring 
below empirically determined thresholds in any single dimension is eliminated. 
This strict intersection mechanism further reduces the corpus from 34.2B 
to 23.2B documents, ensuring that only documents meeting all quality 
requirements are retained for pre-training.

\subsection{Hybrid Deduplication} 
\label{sec:hybrid_deduplication}

Massive repetitive web content not only causes models to overfit templatized structures during training but also systematically compresses the effective learning space for long-tail technical knowledge. To address this, we design a two-stage hybrid deduplication pipeline—where \textit{hybrid} denotes the combination of n-gram-based fuzzy deduplication and embedding-based semantic deduplication—targeting literal and semantic redundancies respectively.

In the first stage, we adopt an $n$-gram-based fuzzy deduplication method employing MinHash-based Locality-Sensitive Hashing (MinHash-LSH). This phase efficiently removes near-duplicate documents with highly overlapping surface text. However, literal text matching has inherent limitations, as it fails to identify semantic redundancies without literal overlap.

In the second stage, we conduct large-scale semantic clustering on the remaining corpus after $n$-gram deduplication. By quantitatively evaluating intra-cluster similarity distributions and performing manual spot checks, we identify two main types of semantic redundancy:

\begin{itemize}[leftmargin=1em, itemindent=0em, itemsep=0.3em]
    \item \textbf{Templatized Clusters}: These documents share an identical structural skeleton and differ only in localized parameter values (e.g., tax rate calculation pages across jurisdictions, state-by-state salary calculators, or color code reference sheets). Because their $n$-gram profiles vary significantly, they easily evade fuzzy deduplication despite being semantically equivalent. Individual high-redundancy clusters can scale up to hundreds of thousands of documents, with intra-cluster duplication rates peaking at $99.96\%$.
    \item \textbf{Redundant Crawling}: This issue arises when the same webpage is crawled at different time points, generating near-duplicate texts with minor content updates or formatting variations that fall below the LSH similarity threshold, allowing them to evade the fuzzy deduplication stage.
\end{itemize}

Existing semantic deduplication frameworks typically apply a fixed, hard threshold (e.g., $0.95$) directly to the cosine similarity of text embeddings for binary hard filtering. We argue that this approach suffers from two fundamental drawbacks. First, the absolute similarity score is highly dependent on the vector space distribution of a specific embedding model\cite{ethayarajh2019contextual}, meaning that identical thresholds map to vastly different degrees of semantic similarity across different models, as empirically observed in our benchmark evaluation. Second, a binary hard threshold inevitably introduces uncontrollable edge-case over-filtering. For instance, documents in specialized technical domains naturally share a dense concentration of domain-specific terminology. Consequently, they tend to cluster near high-similarity thresholds; triggering an immediate deletion based on a single high-similarity match systematically purges these high-value, long-tail technical resources.

To mitigate these challenges, we introduce systematic improvements at both the embedding model selection and deduplication algorithm levels.

\paragraph{Embedding Model Selection} We construct a similarity-interval benchmark annotated by GPT-4o, spanning the $[0.90, 1.00]$ range with a step size of $0.01$, to evaluate four candidate models. Our empirical assessment reveals that jina-v3 \cite{sturua2024jina} suffers from a severe prefix bias, yielding a false positive rate exceeding $40\%$ in high-similarity intervals. The Qwen3 \cite{yang2025qwen3} series (including the $0.6\text{B}$ and $4\text{B}$ variants) falls short in overall discriminative performance. Conversely, embeddinggemma-300m\cite{team2024gemma} demonstrates optimal linear monotonicity and discrimination accuracy across all similarity intervals, and is therefore selected as our production backbone.

\begin{table}[th]
\centering
\caption{Per-interval false positive rates before and after soft deduplication ($T_{\text{target}}= 10$).}
\label{tab:soft_dedup}
\begin{tabular}{lccc}
\toprule
\textbf{Similarity Interval} & \textbf{Weight $w_k$} & \textbf{Baseline (Edge-level) FPR} & \textbf{Ours (Multi-path) FPR} \\
\midrule
$[0.99,\ 1.00]$ & 8.0 & 37.5\% & \textbf{28.03\%} \\
$[0.98,\ 0.99)$ & 7.0 & 34.8\% & \textbf{21.89\%} \\
$[0.96,\ 0.98)$ & 6.0 & 42.2\% & 23.13\% \\
$[0.95,\ 0.96)$ & 5.0 & 54.5\% & 29.70\% \\
$[0.94,\ 0.95)$ & 4.0 & 65.8\% & 35.49\% \\
$[0.92,\ 0.94)$ & 3.0 & 71.6\% & 34.45\% \\
$[0.90,\ 0.92)$ & 2.0 & 85.0\% & 44.37\% \\
\bottomrule
\end{tabular}
\end{table}

\paragraph{Soft Deduplication Algorithm} To mitigate over-filtering at the similarity boundary (cosine score $[0.90, 1.00]$), we propose a \textit{soft deduplication} strategy based on a weighted voting mechanism (details see in Appendix~\ref{subsec:Soft Deduplication Algorithm}). In contrast to conventional hard-threshold pipelines that discard documents immediately after a single high-similarity match, \textit{soft deduplication} accumulates penalty scores from multiple similarity edges within each cluster. 
A document is classified as redundant only if its aggregated score surpasses a predefined \textit{Target Score} (set to 10 in our experiments). 
Each similarity edge is assigned a weighted penalty according to its similarity interval—empirically determined based on observed false positive rates—\textit{e.g.}, $7$ points for $[0.98, 0.99)$ and $8$ points for $[0.99, 1.00]$. 
By requiring structural confirmation across multiple overlapping similarity edges ($N$) within the cluster graph, the joint error propagation is strictly bounded. This multi-path voting mechanism ensures that the collective false positive rate decays substantially compared to single-edge decision boundaries, as the aggregated penalty mitigates localized model estimation noise.

Empirical results demonstrate that \textit{soft deduplication} effectively reduces over-filtering risk: the false positive rate drops from 37.5\% to 28.03\% in the $[0.99, 1.00]$ interval, and from 34.8\% to 21.89\% in the $[0.98, 0.99)$ interval. Table~\ref{tab:soft_dedup} reports the complete per-interval false positive rates before and after applying soft deduplication, along with the empirically determined interval weights used in the scoring configuration. At the billion-document scale, this reduction corresponds to the preservation of a substantial volume of additional high-value technical documents.

\subsection{Data Sampling}
\label{sec:data_sampling}
The core goal of our sampling strategy is to construct a high-quality 
and diverse data subset from 6.4B deduplicated documents within 
a fixed token budget.
Based on the \textit{SampleMix} framework~\cite{xi2025samplemix}, we optimize 
it from three aspects: diversity scoring, quality scoring, 
and sampling distribution design.
Specifically, we design a diversity scoring module, a content value-centric quality scoring paradigm, and a power sampling strategy, 
which are jointly applied to construct the final training subset.

\paragraph{Diversity Scoring}
All documents are first mapped into latent embedding space and grouped 
via K-Means clustering ($K \approx 49{,}000$).
For each document $x_i$ belonging to cluster $C_j$, its diversity score 
$d_i$ is defined as the product of intra-cluster compactness and 
inter-cluster separation:
\begin{equation}
    d_i = d_{\text{intra},j} \times d_{\text{inter},j}
\end{equation}
A lower intra-cluster compactness score penalizes overly dense clusters, 
discouraging oversampling of redundant content, while a higher 
inter-cluster separation encourages sampling from semantically distinct 
regions of the embedding space.

\paragraph{Content Value-Centric Quality Scoring}
We observe an inherent drawback of the linear summation strategy adopted 
in \textit{SampleMix}: it fails to balance superficial presentation and content 
value, tending to retain verbose yet logically flawed content while 
eliminating insightful technical texts with minor presentation defects.
To address this, we propose a value-centric scoring paradigm that bypasses superficial presentation and directly targets core information value.

Specifically, we define the content value $q_{i}$ by aggregating value-oriented dimensions—including \textit{knowledge\_score}, \textit{education\_score}, \textit{helpfulness\_score}, and \textit{reasoning\_score} via linear summation. This reward-oriented mechanism assigns higher priority to documents with outstanding professional merits, serving as our primary quality metric for downstream sampling. 
Before power sampling, documents failing to meet the value threshold ($q_i \leq 1$) are filtered out based on the aggregated content-value score.

\paragraph{Power Sampling}
\label{sec:power_sampling}
In the final data screening stage, we fuse the normalized quality score 
and diversity score into a unified sampling weight:
\begin{equation}
    s_i = \alpha \cdot q_i + (1-\alpha) \cdot d_i
\end{equation}
where $\alpha = 0.6$ balances quality and diversity.
Different from the Softmax-based sampling used in \textit{SampleMix}, 
we adopt Power Sampling to sharpen the probability distribution:
\begin{equation}
p_i = \frac{\tilde{x}_i^{\,n} \cdot \sigma\!\left(k(\tilde{x}_i - a)\right)}{\displaystyle\sum_j \tilde{x}_j^{\,n} \cdot
  \sigma\!\left(k(\tilde{x}_j - a)\right)}
\end{equation}
where $\tilde{x}_i \in [0, 1]$ is the min-max normalized score of document $s_i$, $n$ is the power exponent that amplifies
the weight gap between high- and low-quality documents, $\sigma(\cdot)$ is the sigmoid function acting as a soft threshold
gate, $k$ controls the steepness of the gate such that a larger $k$ produces a sharper cutoff, and $a$ is the center point of
the gate below which documents are progressively suppressed.
Compared with Softmax, Power Sampling exhibits sharper score 
differentiation in the high-score interval, realizing targeted 
oversampling of high-quality samples while suppressing low-quality ones.

\section{Experiments}
\subsection{Experimental Setup}

\paragraph{Model Architecture and Training Setup}
To rigorously evaluate the effectiveness of \textsc{CuraWeb}, we train all models from scratch with a parameter scale of 3B. 
To isolate the pre-training corpus as the sole experimental variable, all models share an identical architecture and the same training hyperparameters, as detailed in Table~\ref{tab:training_config}. 
Our primary evaluation is conducted under a training budget of 200B tokens. Furthermore, we extend the two best-performing configurations at the 200B scale to 1T tokens to analyze their scaling behavior across different training corpora.

\paragraph{Baselines and Evaluation}
We benchmark our approach against several widely adopted, high-quality public pre-training corpora under the same token budget, including DCLM~\cite{DCLM-2024}, FineWeb-Edu~\cite{penedo2024fineweb}, Nemotron-CC~\cite{Nemotron-CC-2025}, and Dolma3~\cite{soldaini2024dolma}. 
Model evaluation is conducted using the \texttt{lm-evaluation-harness} framework~\cite{eval-harness}, spanning multiple capability dimensions such as mathematical and logical reasoning, multidisciplinary knowledge, reading comprehension and commonsense reasoning. 
In total, 10 classic benchmarks are included: GSM8K-Platinum~\cite{cobbe2021training}, MathQA~\cite{amini2019mathqa}, MMLU (5-shot)~\cite{hendrycks2021measuring}, MMLU-Pro (5-shot)~\cite{wang2024mmlu}, RACE~\cite{lai2017race}, HellaSwag~\cite{zellers2019hellaswag}, PIQA~\cite{bisk2020piqa}, SCIQ~\cite{welbl2017crowdsourcing}, Winogrande~\cite{sakaguchi2021winogrande}, and OpenBookQA~\cite{mihaylov2018can}. We report both the individual accuracy for each task and the unweighted macro-average score across all 10 benchmarks.

\subsection{Main Results}

\begin{figure}[th]
  \centering

  \begin{subfigure}[t]{0.48\textwidth}
    \centering
    \includegraphics[width=\linewidth]{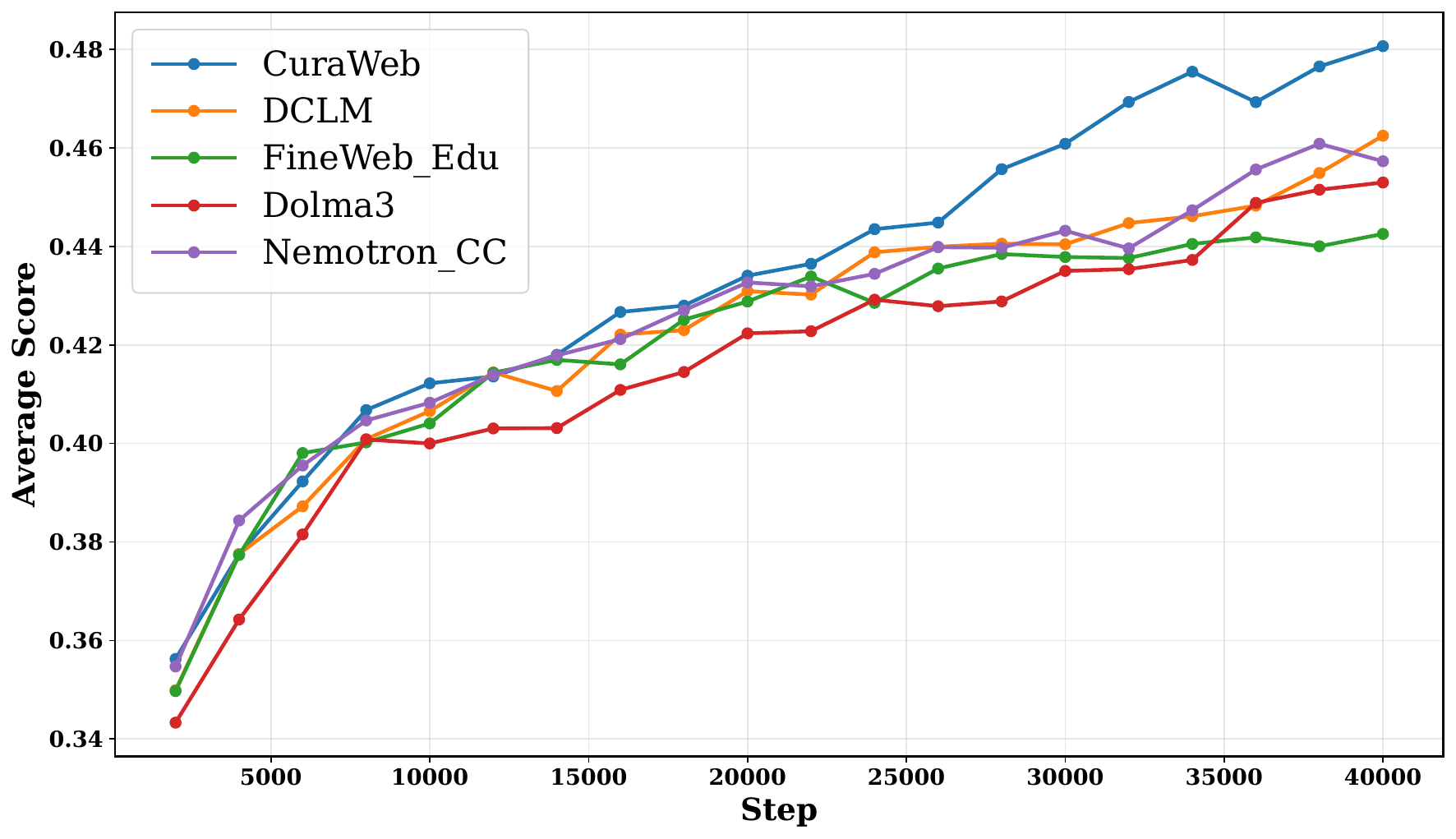}
    \caption{Downstream performance at 200B tokens.}
    \label{fig:baseline_comparison}
  \end{subfigure}
  \hfill
  \begin{subfigure}[t]{0.48\textwidth}
    \centering
    \includegraphics[width=\linewidth]{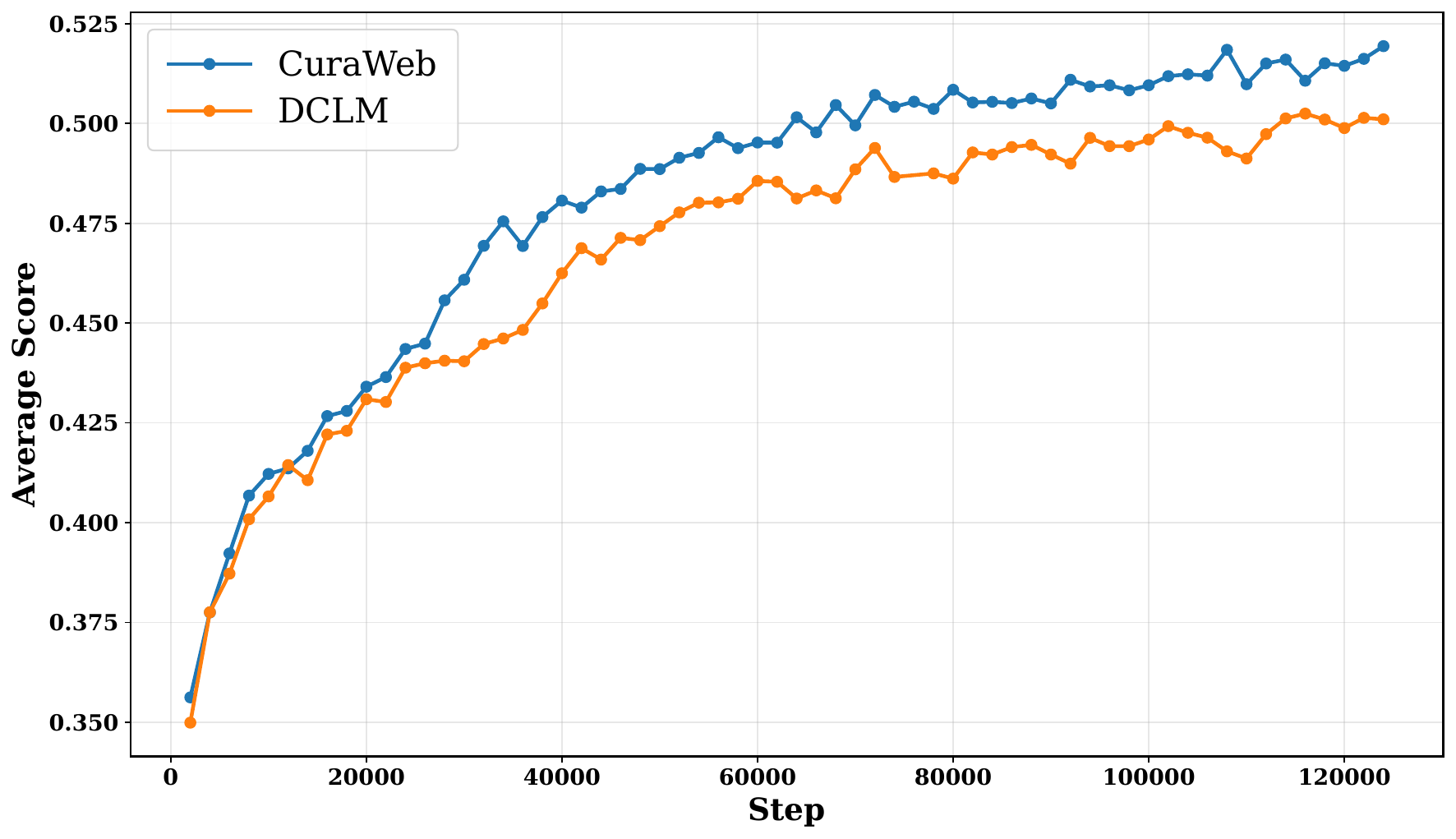}
    \caption{Scaling trends up to 1T tokens.}
    \label{fig:scaling}
  \end{subfigure}

  \caption{
    Evaluation results and scaling behaviors. 
    (a) Comparison between \textsc{CuraWeb} and alternative baseline corpora under a strict 200B token budget. 
    (b) Performance margins between \textsc{CuraWeb} and DCLM as training scales from 200B to 1T tokens.
  }
  \label{fig:main_and_scaling}
\end{figure}

Table~\ref{tab:main_results} and Figure~\ref{fig:baseline_comparison} present a comprehensive evaluation. Under a strict 200B token budget, \textsc{CuraWeb} achieves an average score of 48.07\%, outperforming the strongest comprehensive baseline (DCLM) by 1.82\% absolute points, and establishing a new state-of-the-art across public corpora. 
Rather than merely trading off performance between specific domains, \textsc{CuraWeb} demonstrates a superior ability to break the "specialization bottlenecks" inherent in existing corpora:

\begin{itemize}[leftmargin=1em, itemindent=0em, itemsep=0.3em]
    \item \textbf{Advantage in Reasoning and Knowledge-Intensive Tasks:} 
    On high-order cognitive benchmarks, \textsc{CuraWeb} demonstrates a clear advantage over existing public corpora. Specifically, it achieves an absolute improvement of 4.39\% on GSM8K compared to the best baseline (Dolma3), while outperforming all baselines on MMLU (+6.61\% over Dolma3) and MMLU-Pro (+2.82\% over Dolma3). This solid growth suggests that our corpus design effectively supports the development of systemic reasoning capabilities.

    \item \textbf{Mitigating Domain Bias in Specialized Baselines:} 
    The results highlight a trade-off in specialized baseline corpora, where top performance in single domains often comes at the expense of general capability. 
    For instance, while FineWeb-Edu performs well on textbook-style data (44.20\% on OBQA), its performance drops significantly on MMLU (28.02\%). Similarly, Nemotron-CC shows strength in certain commonsense tasks but lags behind in mathematical reasoning. In contrast, \textsc{CuraWeb} maintains balanced performance, ranking as either the best or second-best across nearly all ten benchmarks.

    \item \textbf{Effectiveness of STEM Data Recovery:} 
    Compared to FineWeb-Edu, \textsc{CuraWeb} yields consistent improvements on STEM-centric benchmarks such as MathQA (+0.47\%) and SCIQ (+0.70\%). This supports our hypothesis that aggressive filtering can inadvertently discard valid scientific texts, and demonstrates that our targeted recall mechanism successfully recovers this knowledge without introducing low-quality noise.
\end{itemize}

\paragraph{Scaling Behavior}
Figure~\ref{fig:scaling} illustrates the performance trends of \textsc{CuraWeb} and DCLM as the training data scales from 200B to 1T tokens.
While \textsc{CuraWeb} outpaces DCLM by 1.82\% at the 200B milestone, it consistently maintains this advantage up to the 1T token mark, preserving a stable 1.95\% performance margin.
This parallel scaling trajectory demonstrates that \textsc{CuraWeb} sustains long-term training efficacy without encountering early performance saturation.

Conversely, baseline corpora that rely strictly on single-dimensional filtering risk premature performance degradation once the token count exceeds their unique data capacity.
Prior studies~\cite{muennighoff2024scaling} have demonstrated that while modest data repetition is tolerable, excessive multi-epoch training yields sharply diminishing returns.
Because \textsc{CuraWeb} preserves its competitive margin throughout scaling, it suggests that its enhanced data diversity successfully mitigates this reduction in training efficiency, thereby validating its suitability for large-scale pre-training.

\begin{table}[th]
\centering
\caption{Downstream task performance evaluated at 200B training tokens. The best results are indicated in bold, and the second-best results are underlined.}
\label{tab:main_results}
\resizebox{\textwidth}{!}{
\begin{tabular}{lcccccccccc|c}
\toprule
\textbf{Dataset} & \textbf{GSM8K} & \textbf{MathQA} & \textbf{MMLU} & \textbf{MMLU-Pro} & \textbf{RACE} & \textbf{HellaSwag} & \textbf{PIQA} & \textbf{SCIQ} & \textbf{Wino.} &
\textbf{OBQA} & \textbf{Avg.} \\
\midrule
DCLM & 3.72 & 24.02 & 39.15 & 11.11 & \textbf{36.94} & \underline{71.63} & 77.37 & 90.30 & \textbf{66.46} & 41.80 & \underline{46.25} \\
FineWeb\_Edu & 3.64 & \underline{25.53} & 28.02 & 9.20 & 35.02 & 67.16 & 76.28 & 90.20 & 63.30 & \textbf{44.20} & 44.26 \\
Nemotron\_CC & 2.56 & 24.96 & 36.25 & 10.04 & 36.75 & \textbf{72.76} & \textbf{78.13} & \textbf{92.10} & 63.46 & 40.30 & 45.73 \\
Dolma3 & \underline{3.80} & 24.92 & \underline{40.56} & \underline{12.08} & 36.65 & 66.94 & 75.84 & 88.70 & 62.90 & 40.60 & 45.30 \\
\midrule
\textbf{CuraWeb} & \textbf{8.19} & \textbf{26.00} & \textbf{47.16} & \textbf{14.90} & \underline{36.84} & 71.09 & \underline{77.86} & \underline{90.90} & \underline{63.85} & \underline{43.90} & \textbf{48.07} \\
\bottomrule
\end{tabular}}
\end{table}

\subsection{Ablation Studies}
\label{sec:ablation}
\begin{figure}[th]
  \centering
  \begin{subfigure}[t]{0.32\textwidth}
    \centering
    \includegraphics[width=\linewidth]{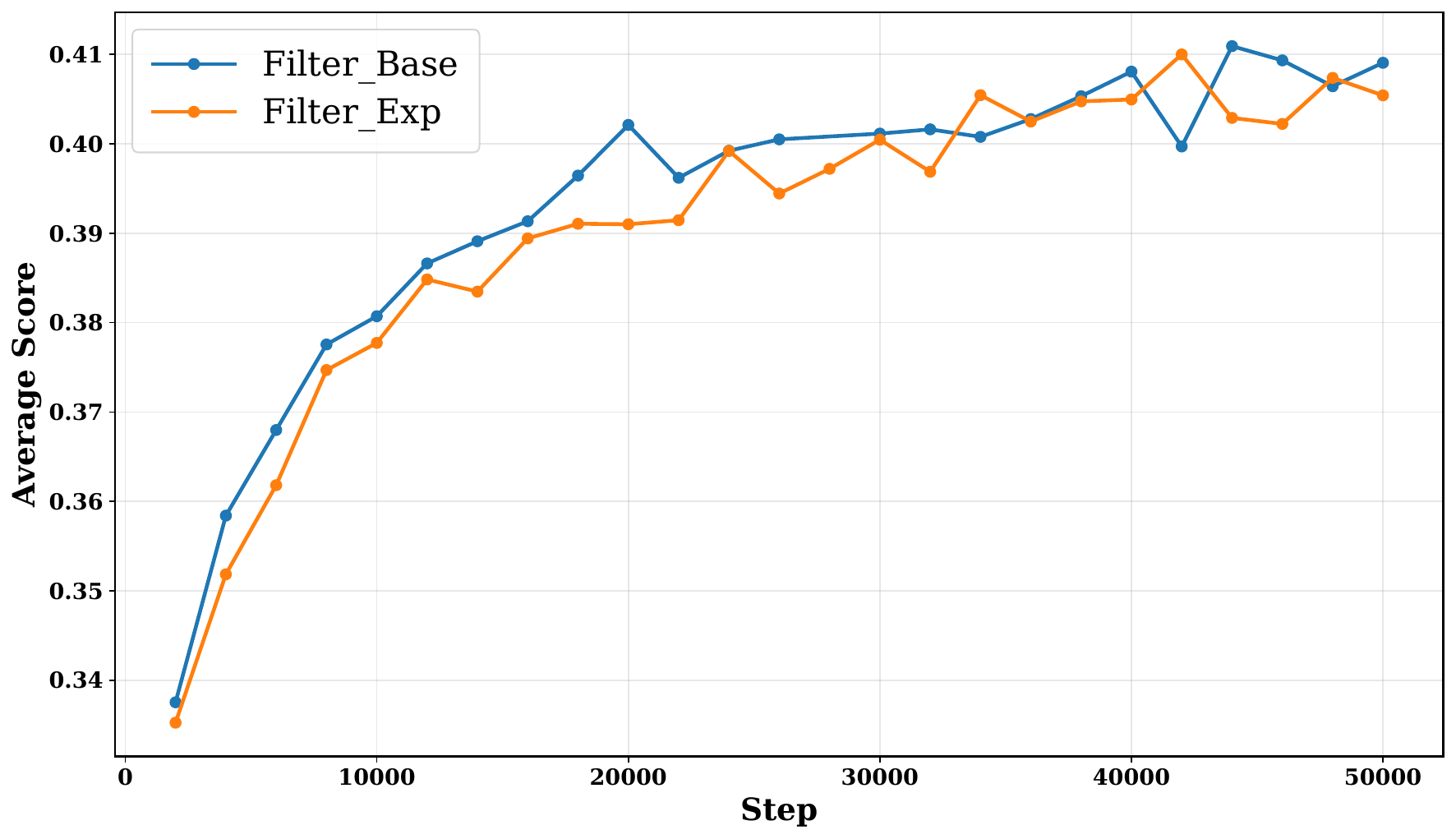}
    \caption{Effect of Data Filtering.}
    \label{fig:ablation_filter}
  \end{subfigure}
  \hfill
  \begin{subfigure}[t]{0.32\textwidth}
    \centering
    \includegraphics[width=\linewidth]{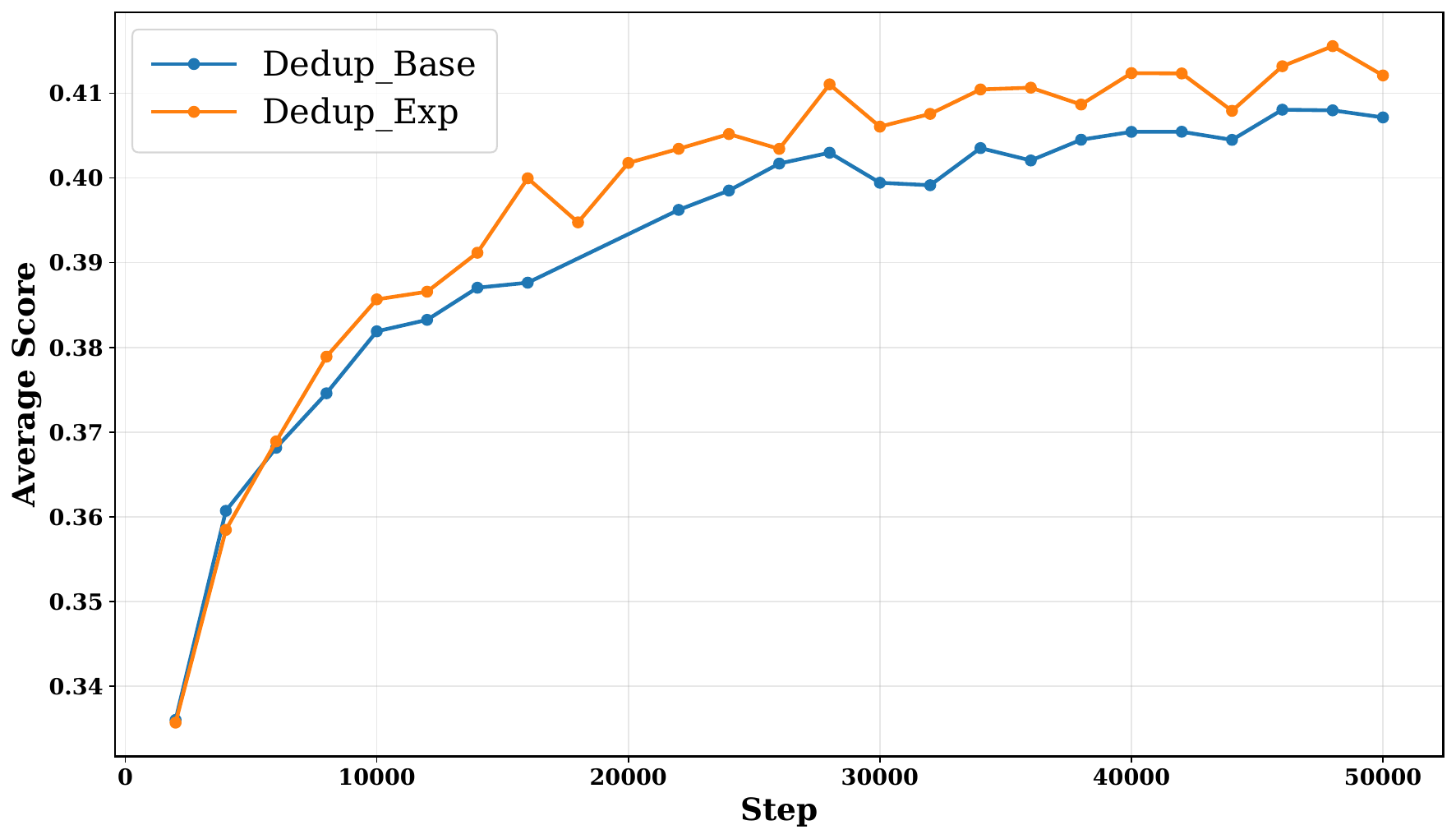}
    \caption{Effect of Deduplication.}
    \label{fig:ablation_dedup}
  \end{subfigure}
  \hfill
  \begin{subfigure}[t]{0.32\textwidth}
    \centering
    \includegraphics[width=\linewidth]{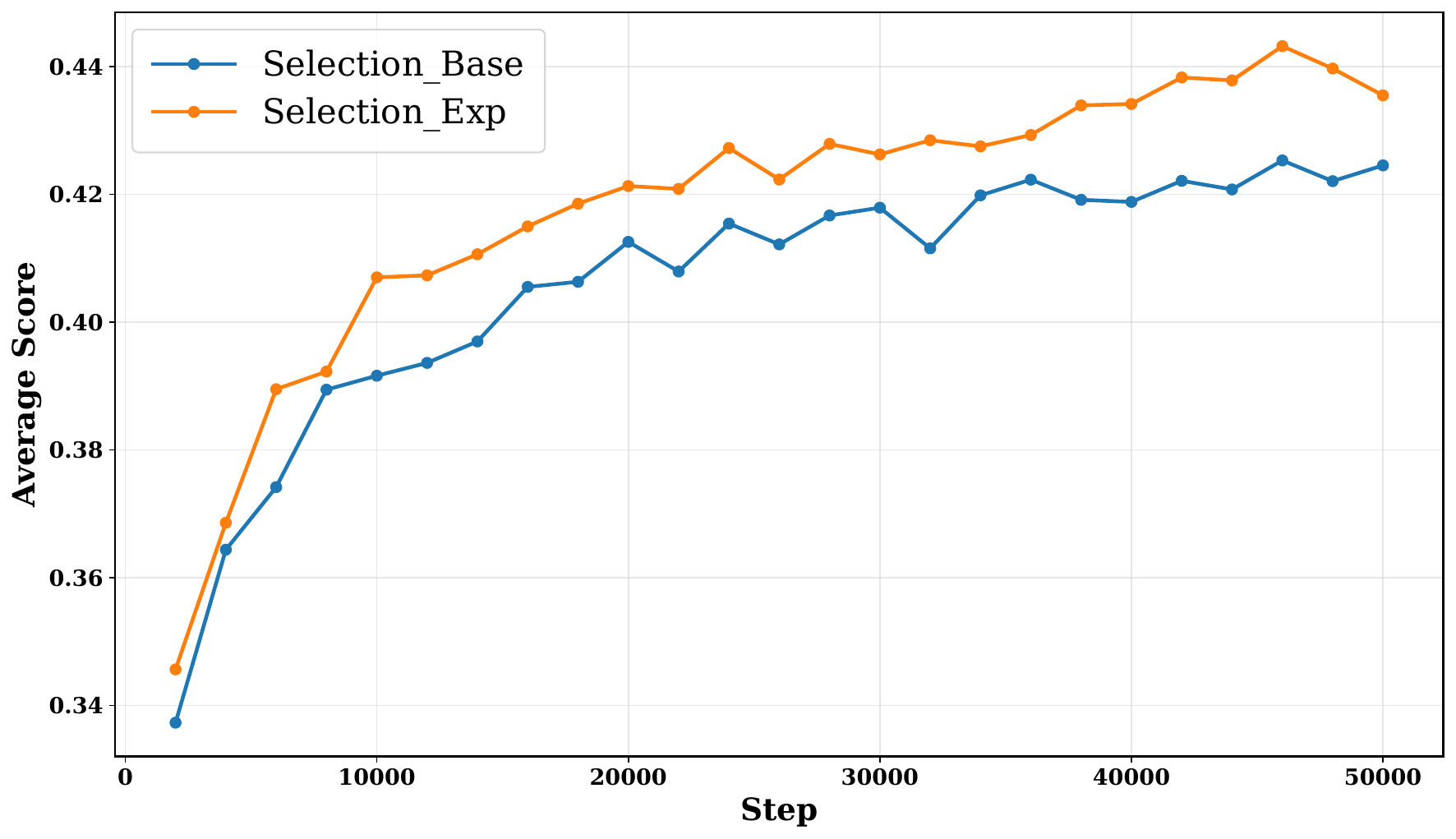}
    \caption{Effect of Sampling Strategy.}
    \label{fig:ablation_sampling}
  \end{subfigure}

\caption{
    Component-wise ablation studies showing downstream performance trends. The performance results are shown for each target component while isolating other variables: 
    (a) Comparing full filtering pipeline vs. purely rule-based filtering. 
    (b) Comparing exact vs. semantic deduplication across token steps. 
    (c) Comparing quality-diversity sampling vs. uniform sampling.
  }
  \label{fig:ablation_all}
\end{figure}

To quantify the independent contribution of each component, we conduct three sets of controlled ablation experiments. 
In each configuration, we isolate and replace only the target component while keeping all other variables constant. 
The resulting downstream performance trends for data filtering, deduplication, and sampling strategies are consolidated in Figure~\ref{fig:ablation_all}(a)--(c), respectively.

The ablation settings demonstrate that each component contributes uniquely to either performance enhancement or data scaling.
As shown in Figure~\ref{fig:ablation_filter}, while removing model-based filtering maintains a similar downstream performance baseline, our full filtering pipeline effectively doubles the volume of retained high-quality data. This confirms that model-based filtering drastically expands data yield without compromising training efficiency (see Section \ref{sec:data_filtering}). 
In contrast, the removal of semantic deduplication (Figure~\ref{fig:ablation_dedup}) shifts the training curve downward, indicating that templated redundancies cause a substantial loss in performance.
Furthermore, replacing our sampling strategy with uniform sampling leads to a visible performance drop (Figure~\ref{fig:ablation_sampling}), validating the necessity of our content value-centric quality scoring and Power Sampling mechanisms.

\section{Analysis}
In this section, we present a systematic evaluation of our data curation pipeline against baseline methods. Our comparative analysis is structured around three core dimensions: (1)~Data Coverage (2)~Data Diversity and (3)~Data Quality.

\subsection{Data Coverage Analysis}
Data coverage evaluates whether the dataset encompasses the comprehensive knowledge 
scope required for foundational model training. Focusing on rule-based and model-based 
filtering---the two stages responsible for the highest data compression---we analyze 
how each filtering stage affects the L1-level domain distribution.

\begin{figure}[h]
  \centering
  \begin{subfigure}[t]{0.32\textwidth}
    \centering
    \includegraphics[width=\linewidth]{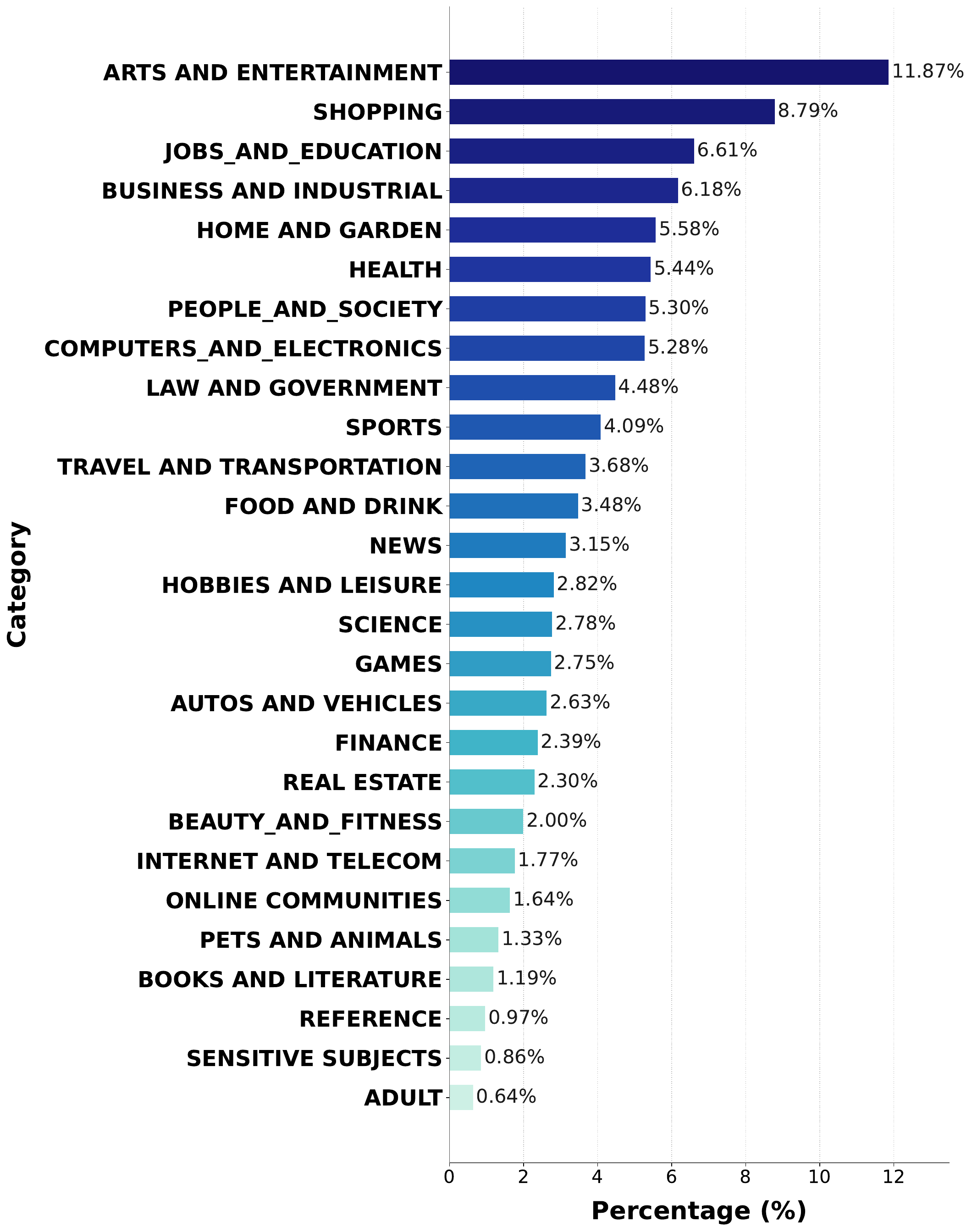}
    \caption{Before filtering.}
    \label{fig:rule_before_l1}
  \end{subfigure}
  \hfill
  \begin{subfigure}[t]{0.32\textwidth}
    \centering
    \includegraphics[width=\linewidth]{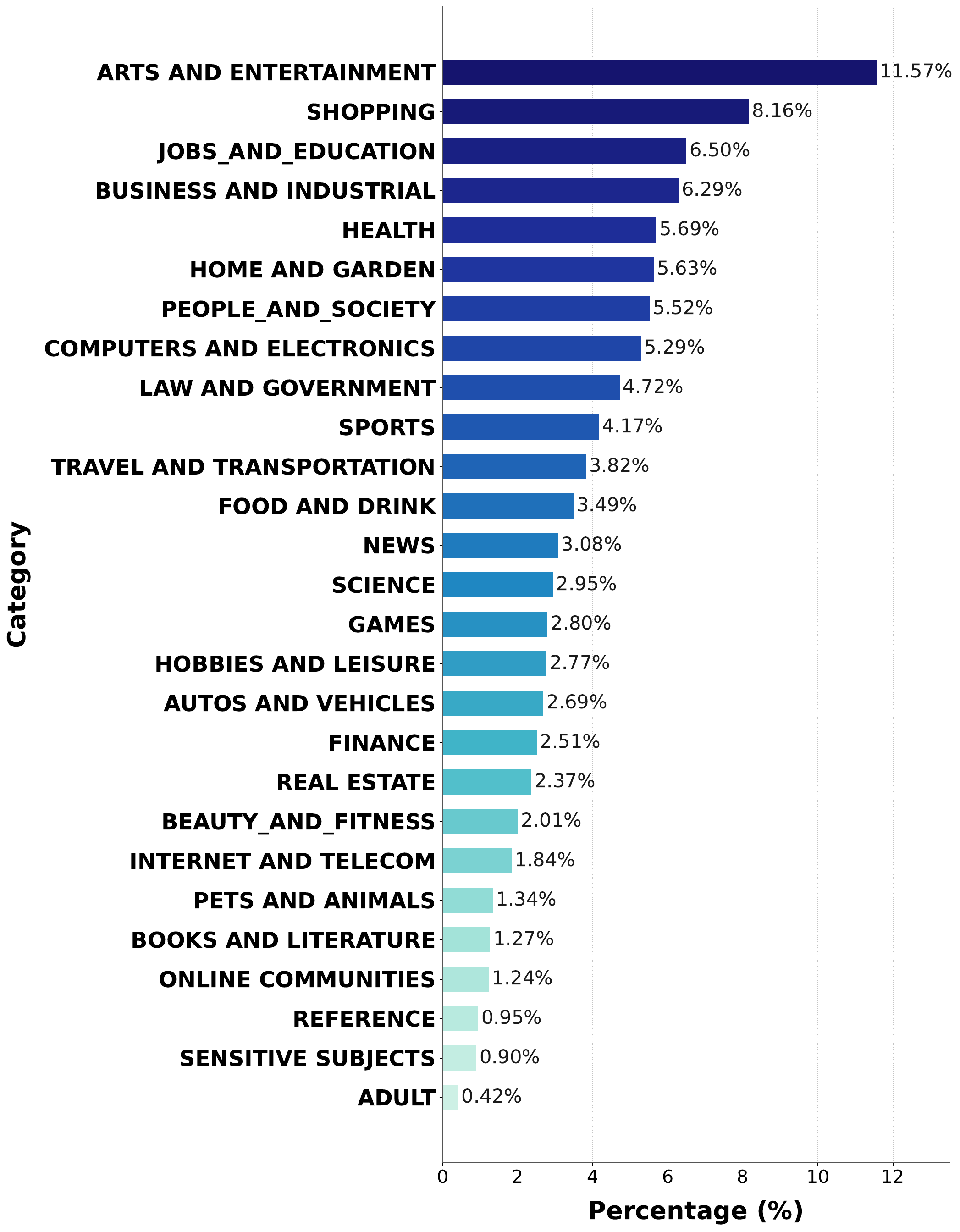}
    \caption{After filtering.}
    \label{fig:rule_after_l1}
  \end{subfigure}
  \hfill
  \begin{subfigure}[t]{0.32\textwidth}
    \centering
    \includegraphics[width=\linewidth]{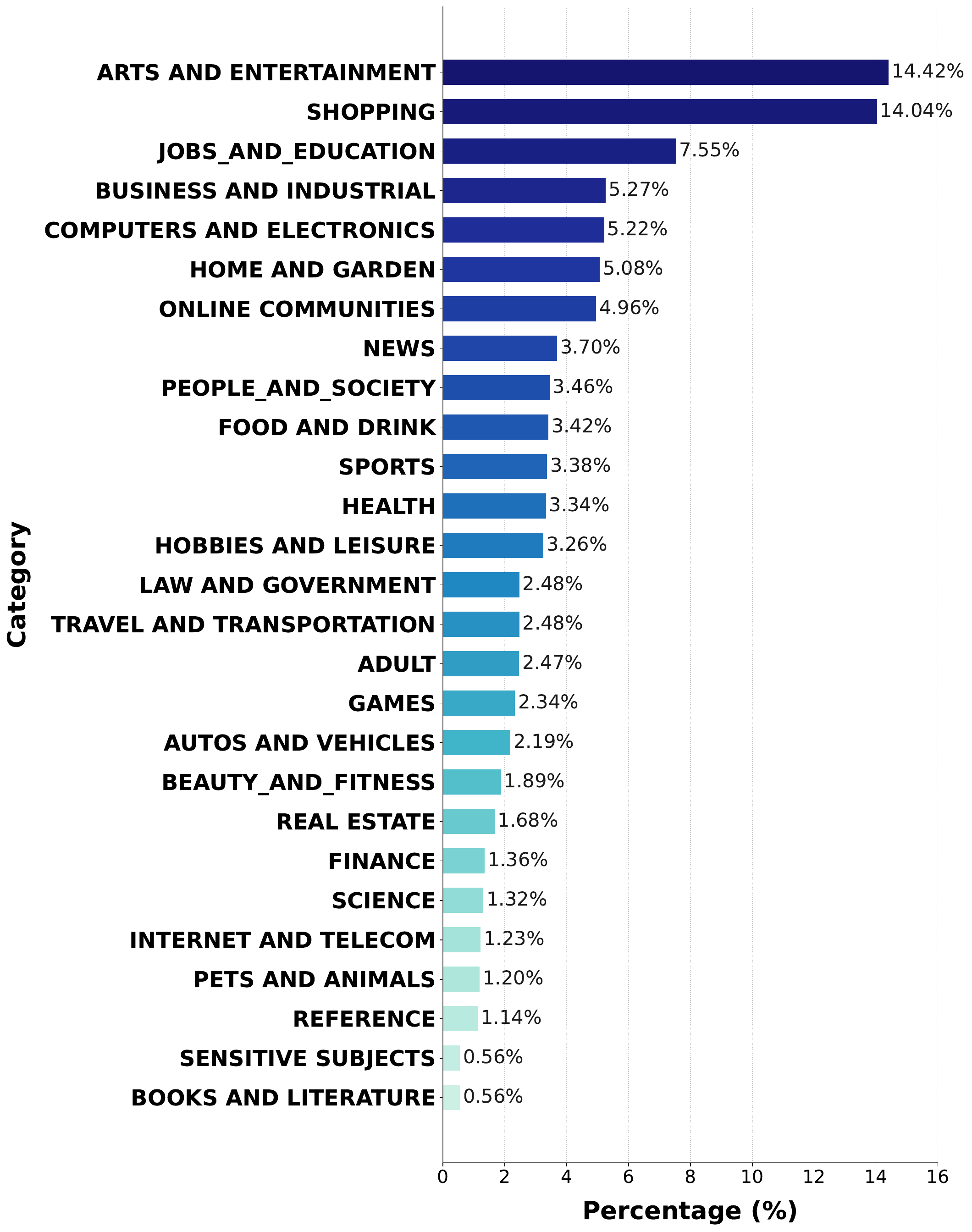}
    \caption{Removed documents.}
    \label{fig:rule_filtered_l1}
  \end{subfigure}
  \caption{
  Domain L1 distribution before and after rule-based filtering:
  (a) before filtering, (b) after filtering, and (c) removed documents.
  }
  \label{fig:rule_filter_l1_all}
\end{figure}

\paragraph{Impact of Rule-Based Filtering on Domain Distribution.}
Figures~\ref{fig:rule_before_l1} and~\ref{fig:rule_after_l1} present the L1-level 
domain distributions before and after rule-based filtering, while 
Figure~\ref{fig:rule_filtered_l1} shows the domain composition of the filtered 
documents. The filtering rate is approximately 10\%, and the resulting 
distributional perturbation is minimal: major categories such as \texttt{Arts}, 
\texttt{Shopping}, and \texttt{Online\_Communities} exhibit only slight decreases of 
0.2\%--0.7\%, and no L1-level domain is significantly compressed. Examining the 
filtered documents, the fastText templated-content classifier primarily targets 
advertisement-related content, whereas the Gopher rules remove documents relatively 
uniformly across domains, consistent with the high-recall design of the rule-based 
filtering stage.

\begin{figure}[h]
  \centering
  \begin{subfigure}[t]{0.48\textwidth}
    \centering
    \includegraphics[height=0.38\textheight, keepaspectratio]{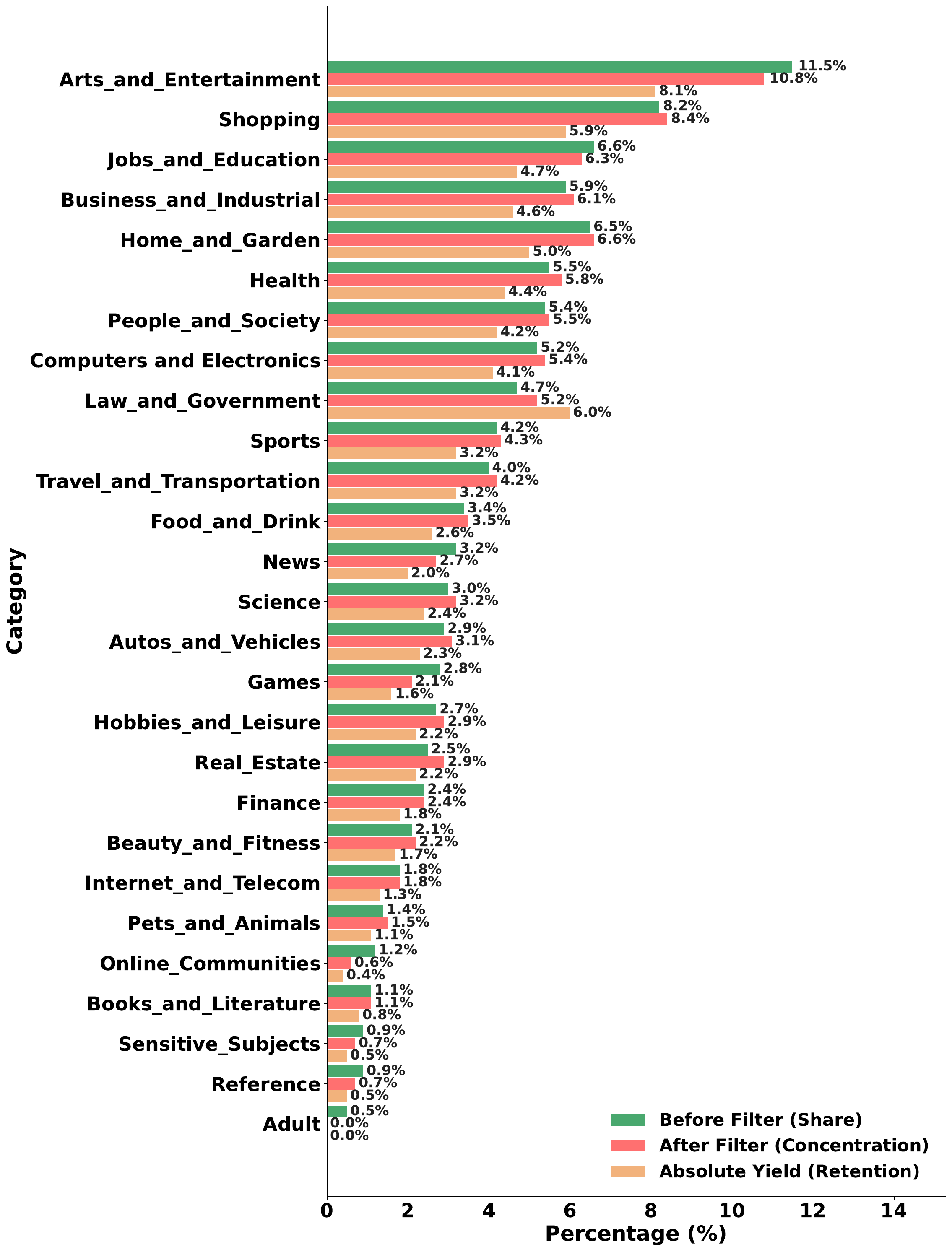}
    \caption{L1-Domain distribution across pipeline stages.}
    \label{fig:model_filter_l1_3bars}
  \end{subfigure}
  \hfill
  \begin{subfigure}[t]{0.48\textwidth}
    \centering
    \includegraphics[height=0.38\textheight, keepaspectratio]{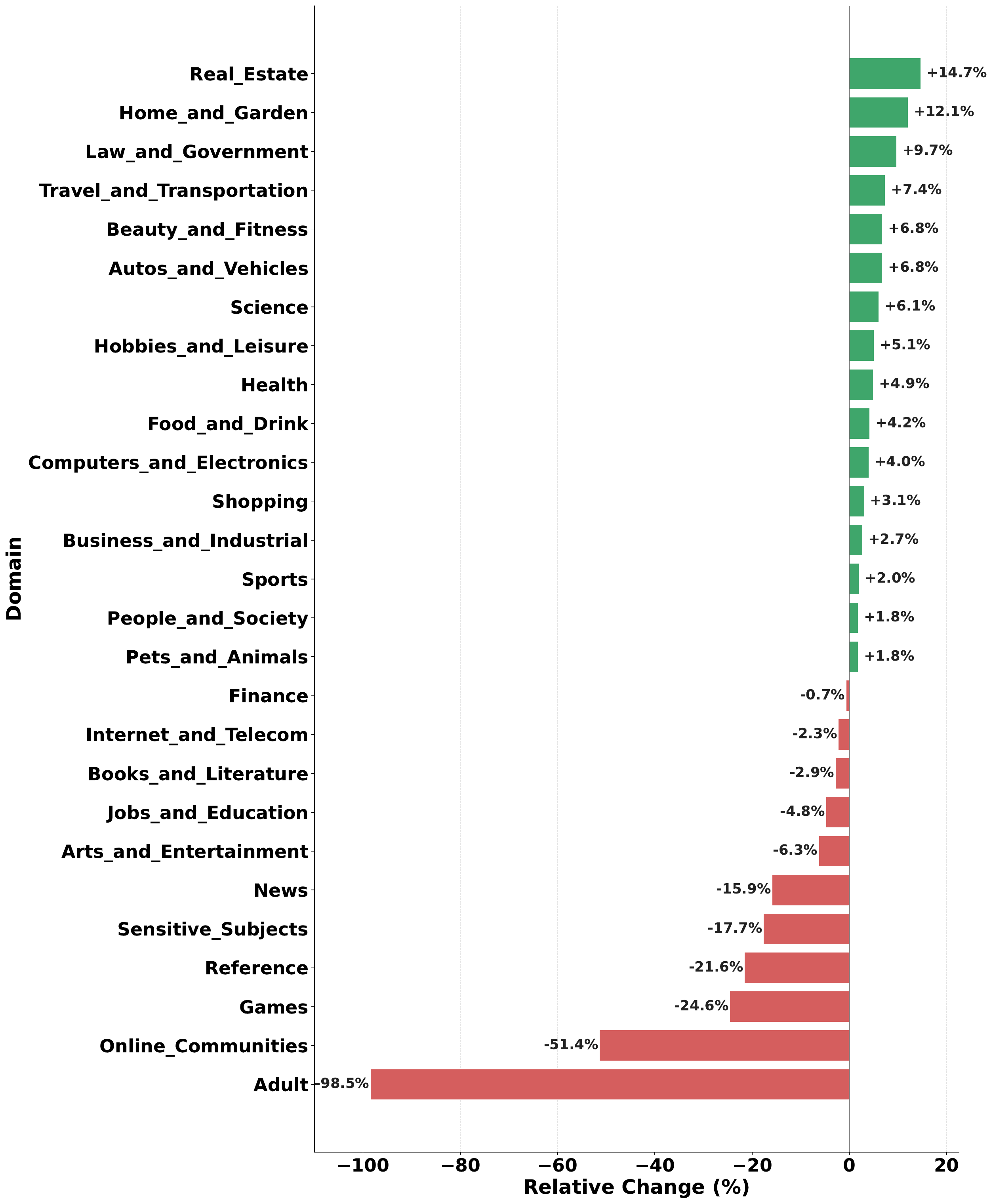}
    \caption{Relative change rate of L1-domain distribution.}
    \label{fig:model_filter_l1_change}
  \end{subfigure}
  \caption{
  Domain-level effects of model-based filtering:
  (a) L1-domain distribution across pipeline stages;
  (b) relative change rate of the domain L1 distribution after model-based filtering.
  }
  \label{fig:model_filter_l1_all}
\end{figure}

\paragraph{Impact of Model-Based Filtering on Domain Distribution.}
Figure~\ref{fig:model_filter_l1_3bars} presents the domain distributions across three 
stages: original, after rule-based filtering, and after model-based filtering. 
Model-based filtering removes an additional 24.8\% of documents, exhibiting selective 
rather than uniform compression: \texttt{Adults}, \texttt{Online\_Communities}, 
\texttt{Games}, and \texttt{Reference} are filtered at substantially higher rates than 
the overall average, while long-tail vertical domains such as \texttt{Religion} and 
\texttt{Apparel} are barely affected. The primary rejection criterion is failure in 
the format dimension---structural deficiencies such as garbled text, sentence 
fragmentation, or excessive noise---rather than issues related to factual correctness or opinion quality.

\paragraph{Comparison of Domain Coverage with Baseline Corpora.}
Under a controlled setting of 50 million documents, we compare the L1-level domain 
distributions of \textsc{CuraWeb} with four mainstream external datasets 
(see Figure~\ref{fig:cross_dataset_domains}). \textsc{CuraWeb} exhibits the most balanced 
distribution: its top three categories (\texttt{Health}, 
\texttt{Arts\_and\_Entertainment}, \texttt{Law\_and\_Government}) together account for 
only 26.5\%. In contrast, FineWeb-Edu's top three categories alone reach 48.2\%, 
reflecting its strong concentration on educational content at the cost of coverage in 
domains such as law, business, and entertainment. Dolma3 shows a technology and 
entertainment bias (\texttt{Computers\_and\_Electronics} and 
\texttt{Arts\_and\_Entertainment} together: 26.4\%), while DCLM and Nemotron-CC 
exhibit similar distribution patterns dominated by \texttt{Arts\_and\_Entertainment}. 
Regarding long-tail domain retention, \textsc{CuraWeb} consistently achieves higher 
proportions in niche yet high-value areas such as 
\texttt{Science/Biological\_Sciences}, \texttt{Finance}, and \texttt{Reference} 
compared with all external datasets.

\subsection{Diversity Analysis}
We quantify the diversity of \textsc{CuraWeb} from three distinct perspectives: \textbf{deduplication redundancy}, \textbf{domain distribution entropy}, and \textbf{lexical diversity}. To ensure a fair evaluation, all comparisons are conducted under a strictly controlled setting of 50M documents per corpus.

\paragraph{Deduplication Redundancy.}
We validate the cascading benefits of our two-stage deduplication pipeline. The original corpus of 23.29B documents is reduced to 7.97B after MinHash fuzzy deduplication (compression ratio: 65.8\%), and further to 6.42B after semantic deduplication (an additional reduction of 19.4\%), yielding a cumulative deduplication rate of 72.4\%.

Semantic deduplication proves particularly effective at capturing near-duplicates that evade MinHash-based fuzzy matching. Specifically, in mathematics and physics corpora, it identifies an additional 682.4K duplicates beyond fuzzy deduplication, bringing the total to 1.65M. Meanwhile, for utility-oriented websites, it expands the duplicate detection coverage from 2.33M to 3.93M documents. Notably, the average Jaccard similarity of the remaining pairs drops from 0.057\% to 0.049\%, confirming the successful elimination of low-similarity templated noise. This dual improvement demonstrates that semantic deduplication substantially broadens duplicate recall without sacrificing precision.

\paragraph{Domain Distribution Entropy.}
Table~\ref{tab:entropy} reports the domain distribution entropy of \textsc{CuraWeb} 
and baseline corpora. \textsc{CuraWeb} achieves a L1-level domain entropy of 4.295 bits 
and a L2-level domain entropy of 6.075 bits, substantially higher than FineWeb-Edu 
(3.733 / 5.477 bits) and Dolma~3 (4.086 / 6.034 bits), while remaining comparable 
to DCLM (4.308 / 6.122 bits) and Nemotron-CC (4.380 / 6.148 bits). To understand 
how this entropy level is achieved and maintained throughout the pipeline, we track 
entropy dynamics across all stages. The results show that filtering and deduplication 
stages introduce negligible entropy loss: rule-filter (4.419) $\rightarrow$ 
model-filter (4.428) $\rightarrow$ fuzzy-dedup (4.427) $\rightarrow$ semantic-dedup 
(4.417) $\rightarrow$ value-filter (4.330) $\rightarrow$ final sampling (4.295). 
The cumulative entropy reduction of approximately 0.13 bits originates primarily from 
the final sampling stage, where a minor entropy reduction is traded for a higher 
proportion of high-value documents and improved downstream performance 
(Table~\ref{tab:main_results}).

\begin{table}[th]
  \centering
    \caption{Domain distribution entropy across corpora (50M documents each), where higher entropy indicates broader domain coverage.}
  \label{tab:entropy}
  \begin{tabular}{lcc}
  \toprule
  \textbf{Dataset} & \textbf{ L1-Doamin Entropy ($H_\text{L1}$)} & \textbf{L2-domain Entropy ($H_\text{L2}$)} \\
  \midrule
  Common Crawl & 4.442 & 5.917 \\
  DCLM                 & 4.308 & 6.122 \\
  Nemotron-CC          & 4.380 & 6.148 \\
  Dolma3               & 4.086 & 6.034 \\
  FineWeb-Edu          & 3.733 & 5.477 \\
  CuraWeb  & 4.295 & 6.075 \\
  \bottomrule
  \end{tabular}
\end{table}

\paragraph{Lexical Diversity.}
We measure lexical diversity using the unique $n$-gram ratio $\mathcal{R}_n$, defined 
as the ratio of unique $n$-grams to total $n$-grams. 
Table~\ref{tab:twr} reports 
$\mathcal{R}_n$ across pipeline stages and baselines. Within the \textsc{CuraWeb} 
pipeline, $\mathcal{R}_n$ increases monotonically from 0.000373 at the rule filtering stage 
to 0.000560 after semantic deduplication, demonstrating that each deduplication stage 
contributes positively to lexical diversity. In the final sampling stage, $\mathcal{R}_n$ 
decreases to 0.000352 due to the density dilution effect of power-law resampling, which 
shifts corpus volume toward high-value long-form documents and increases total $n$-gram 
count from 8.5B to 14.6B. Importantly, the number of unique $n$-grams continues 
to increase to 5.13M---the highest among all compared baselines---indicating that the 
sampled corpus provides broader lexical coverage than earlier pipeline stages.
Compared with baselines, FineWeb-Edu achieves the lowest $\mathcal{R}_n$ (0.000213) 
and unique $n$-gram count (3.18M), consistent with its narrow domain distribution. 
Although Dolma3 achieves the largest unique $n$-gram count (5.67M), its overall 
quality score is substantially lower. When considering entropy, $\mathcal{R}_n$, and 
downstream performance jointly, \textsc{CuraWeb} consistently ranks within the top 
tier across all diversity metrics without sacrificing quality.

\begin{table}[th]
  \centering
\caption{Lexical diversity ($\mathcal{R}_n$) across pipeline stages and baselines (50M documents each).}
  \label{tab:twr}
  \begin{tabular}{lrrrr}
  \toprule
  \textbf{Stage / Dataset} & \textbf{Documents} & \textbf{Total n-grams} & \textbf{Unique n-grams} & \textbf{$\mathcal{R}_n$} \\
  \midrule
  \multicolumn{5}{l}{\emph{CuraWeb pipeline stages}} \\
  Rule Filter             & 50{,}000{,}000 & 12{,}548{,}809{,}607 & 4{,}677{,}970 & 0.000373 \\
  Model Filter                & 50{,}000{,}000 &  9{,}898{,}961{,}669 & 4{,}662{,}598 & 0.000471 \\
  Fuzzy Dedup                 & 50{,}000{,}000 &  9{,}140{,}074{,}677 & 4{,}893{,}923 & 0.000535 \\
  Semantic Dedup              & 50{,}000{,}000 &  8{,}455{,}364{,}183 & 4{,}733{,}528 & 0.000560 \\
  Value Filter                & 50{,}000{,}000 & 12{,}872{,}152{,}074 & 4{,}815{,}215 & 0.000374 \\
  Final & 50{,}000{,}000 & 14{,}573{,}357{,}160 & 5{,}129{,}070 & 0.000352 \\
  \midrule
  \multicolumn{5}{l}{\emph{Baselines}} \\
  Nemotron-CC                 & 50{,}000{,}000 & 11{,}659{,}696{,}700 & 4{,}101{,}843 & 0.000352 \\
  FineWeb-Edu                 & 50{,}000{,}000 & 14{,}944{,}623{,}936 & 3{,}179{,}844 & 0.000213 \\
  Dolma3                     & 50{,}000{,}000 & 17{,}486{,}244{,}342 & 5{,}665{,}350 & 0.000324 \\
  DCLM                        & 50{,}000{,}000 & 15{,}554{,}026{,}736 & 4{,}677{,}300 & 0.000301 \\
  \bottomrule
  \end{tabular}
\end{table}

\subsection{Quality Analysis} 
To systematically evaluate the quality of \textsc{CuraWeb}, we analyze our 
data curation pipeline from four perspectives: 
(1) \textbf{corpus quality assessment}, how the overall quality of our final corpus 
compares with existing open-source datasets across multiple quality dimensions; 
(2) \textbf{filtering precision}, whether our hybrid rule+model strategy accurately 
removes low-quality content without over-discarding high-value documents; 
(3) \textbf{high-value content retention}, how well our pipeline preserves 
information-dense documents across key STEM domains compared with existing corpora; and 
(4) \textbf{sampling-driven quality optimization}, how importance sampling further 
reshapes the corpus distribution toward higher knowledge density.

\paragraph{Corpus Quality Assessment.}
  We evaluate overall corpus quality from two complementary angles.
  Within our pipeline, the overall quality score increases monotonically
  across stages: from 0.300 after fuzzy deduplication, through 0.302 after
  semantic deduplication and 0.436 after value filtering, finally reaching
  0.521 after importance sampling---the highest among all internal pipeline
  variants, indicating that each stage contributes positively to data quality.
  Compared with external datasets, our final corpus (0.521) outperforms
  Dolma3 (0.457), DCLM (0.448), and Nemotron-CC (0.392).
  As shown in Table~\ref{tab:value_score}, \textsc{CuraWeb} outperforms
  all external datasets in Knowledge (1.164) and PracticalHelpfulness (0.751).
  Although FineWeb-Edu achieves a higher overall score (0.610), this advantage
  stems from its strong concentration on educational content, at the cost of
  substantial domain diversity loss ($H_{\text{L1}}=3.73$, the lowest among
  all compared datasets) and a notably lower PracticalHelpfulness score (0.562).
  Overall, \textsc{CuraWeb} achieves the most balanced quality profile
  across all dimensions.

\begin{table}[t]
  \centering
  \caption{Normalized value scores of different corpora across value-oriented evaluation 
  dimensions. The best results are indicated in bold, and the second-best results are underlined.}
  \label{tab:value_score}
  \resizebox{\columnwidth}{!}{
  \begin{tabular}{lccccc}
  \toprule
  \textbf{Dataset} & \textbf{EducationValue} & \textbf{Knowledge} & 
  \textbf{PracticalHelpfulness} & \textbf{Reasoning} & \textbf{Composite\_score}\\
  \midrule
  \multicolumn{6}{l}{\emph{CuraWeb pipeline stages}} \\
  Rule Filter & 1.110 & 0.675 & 0.438 & 0.512 & 0.380 \\
  Model Filter & 0.894 & 0.400 & 0.293 & 0.322 & 0.313 \\
  Fuzzy Dedup & 0.859 & 0.347 & 0.252 & 0.289 & 0.300 \\
  Semantic Dedup & 0.846 & 0.352 & 0.247 & 0.305 & 0.302 \\
  Value Filter & 1.249 & 0.942 & 0.488 & 0.797 & 0.436 \\
  Final        & \underline{1.498} & \underline{1.164} & \textbf{0.751}  & \underline{0.987} & \underline{0.521} \\
  \midrule
  \multicolumn{6}{l}{\emph{Baselines}} \\
  FineWeb-Edu       & \textbf{1.963} & \textbf{1.387} & \underline{0.562} 
  & \textbf{1.202} & \textbf{0.610} \\
  Dolma3            & 1.233 & 0.983 & 0.527 & 0.923 & 0.457 \\
  DCLM              & 1.230 & 0.887 & 0.486 & 0.936 & 0.448 \\
  Nemotron-CC       & 1.114 & 0.684 & 0.398 & 0.640 & 0.392 \\
  \bottomrule
  \end{tabular}
  }
\end{table}

\paragraph{Filtering Precision.}
We compare two filtering strategies in terms of document retention and content quality. 
The pure rule-based strategy retains 34.18B documents, while the rule+model 
hybrid strategy further compresses the corpus to 23.29B documents, removing an 
additional 31.9\% of documents. 
However, pure rule-based methods such as Gopher rely on surface-level textual 
features---including punctuation density, special character ratios, and line-length 
distributions---to assess quality, leading to systematic misclassification of STEM 
documents containing mathematical formulas, chemical symbols, or code snippets. 
In contrast, our rule+model hybrid strategy leverages model-based signals to identify 
semantically low-quality content, effectively distinguishing between ``symbol-dense but 
high-quality'' and genuinely low-quality documents. 
As shown in Table~\ref{tab:stem_retention}, the proportion of STEM documents increases 
from 19.74\% at the rule filtering stage to 33.58\% after importance sampling, while 
math-related documents rise from 1.656\% to 3.832\%, demonstrating that our hybrid 
strategy accurately removes low-quality content while actively protecting high-value 
professional documents.

\paragraph{High-Value Content Retention.}
Beyond filtering precision, we further evaluate how well our pipeline retains 
high-value documents across key STEM domains. 
As reported in Table~\ref{tab:domain_hq_retention}, \textsc{CuraWeb} achieves an 
average retention rate of 0.432 for high-quality documents (score $\geq 2$ under a 
3-point scoring system) across Computer Science, Programming, Mathematics, and Physics, 
consistently outperforming DCLM (0.364), Dolma3 (0.376), and Nemotron-CC (0.261), 
while remaining comparable to FineWeb-Edu (0.427). 
Notably, although FineWeb-Edu achieves a slightly higher retention rate in Programming 
(0.451), its overall domain diversity is substantially lower ($H_{\text{L1}}=3.73$, 
the lowest among all compared datasets). 
\textsc{CuraWeb} thus maintains competitive STEM retention rates while preserving 
a significantly more balanced domain distribution.

\begin{table}[th]
  \centering
  \caption{Retention ratio of code, math and STEM documents across pipeline stages.}
  \label{tab:stem_retention}
  \begin{tabular}{lcccc}
  \toprule
  \textbf{Stage} & \textbf{Code Ratio} & \textbf{Math Ratio} & \textbf{STEM Ratio}  \\
  \midrule
  Rule Filter  & 0.886\% & 1.656\% & 19.74\% \\
  Model Filter     & 0.801\% & 1.701\% & 20.38\% \\
  Fuzzy Dedup      & 0.915\% & 1.789\%  & 25.11\% \\
  Semantic Dedup   & 0.847\% & 1.546\%  & 19.49\% \\
  Value Filtering     & 1.600\% & 2.713\%  & 30.46\% \\
  Final (w/ importance sampling) & \textbf{2.544\%} & \textbf{3.832\%} & \textbf{33.58\%} \\
  \bottomrule
  \end{tabular}
\end{table}

\begin{table}[th]
  \centering
  \caption{High-quality document retention rate ($\geq$2 out of 3) across key STEM 
  domains.}
  \label{tab:domain_hq_retention}
  \begin{tabular}{lccccc}
  \toprule
  \textbf{Dataset} & \textbf{Computer Science} & \textbf{Programming} & 
  \textbf{Mathematics} & \textbf{Physics} & \textbf{Avg} \\
  \midrule
  FineWeb-Edu       & 0.429 & \textbf{0.451} & 0.405 & \textbf{0.424} & 0.427 \\
  Dolma3            & 0.350 & 0.332 & 0.426 & 0.396 & 0.376 \\
  DCLM              & 0.354 & 0.336 & 0.407 & 0.358 & 0.364 \\
  Nemotron-CC       & 0.252 & 0.208 & 0.301 & 0.282 & 0.261 \\
  CuraWeb     & \textbf{0.436} & 0.425 & \textbf{0.448} & 0.419
  & \textbf{0.432} \\
  \bottomrule
  \end{tabular}
\end{table}

\begin{figure}[h]
  \centering
  \includegraphics[width=0.85\textwidth]{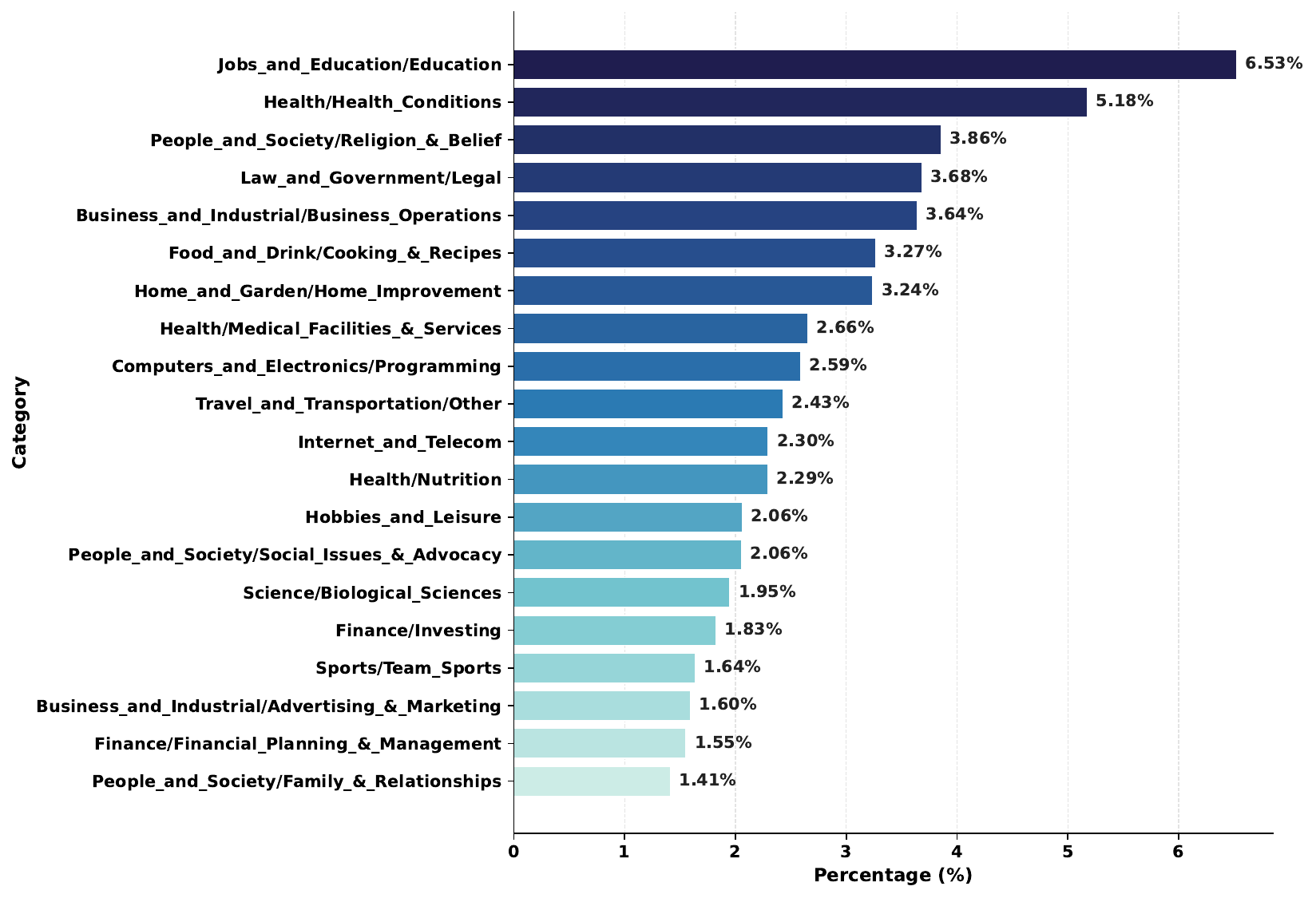}
  \caption{
      Top-20 domain categories by upsampling ratio in the \textsc{CuraWeb} 
      importance sampling stage. Categories with high value density 
      such as \texttt{Education}, \texttt{Health}, and \texttt{Programming} receive the highest upsampling 
      weights.
  }
  \label{fig:upsampled_domain}
\end{figure}

\paragraph{Sampling-Driven Quality Optimization.}
The importance sampling stage clusters documents and adjusts per-cluster 
sampling weights according to value signals, enabling targeted enhancement of 
high quality content. 
As shown in Figure~\ref{fig:upsampled_domain}, the most upsampled domains are 
concentrated in knowledge-intensive categories, including 
\texttt{Jobs\_and\_Education/Education} (6.53\%), \texttt{Health/Health\_Conditions} 
(5.18\%), \texttt{Law\_and\_Government/Legal} (3.68\%), and 
\texttt{Computers\_and\_Electronics/Programming} (2.59\%), while 
low-information-density categories such as \texttt{Shopping} and \texttt{Real\_Estate} 
experience substantial declines. 
This is further reflected in the overall domain distribution: after sampling, 
\texttt{Health} rises to the largest category (9.11\%), and the proportions of 
\texttt{Science} (7.83\%) and \texttt{Jobs\_and\_Education} (7.04\%) increase 
significantly, whereas \texttt{Arts\_and\_Entertainment} drops from the top position 
(9.32\%) before sampling. 
These structural shifts confirm that importance sampling effectively reallocates corpus weights toward domains with higher value density, consistent with our design objective. At the lexical level, the sampled corpus contains 5.13M unique n-grams and 14.6B total tokens, representing a substantial scale-up compared to the semantic deduplication stage (4.73M unique n-grams and 8.5B total tokens), while sustaining a high lexical diversity ($\mathcal{R}_n = 0.000352$). 
This trajectory indicates that the sampling stage successfully enriches the corpus with diversity and information-dense documents, thereby improving overall data quality without inducing lexical dilution or redundancy.

\section{Conclusion}
In this paper, we introduced \textsc{CuraWeb}, a large-scale pre-training dataset built using a unified data cleaning pipeline. Instead of using isolated and simple filtering steps, our approach relies on multi-dimensional, signal-driven data management. To balance data quality and diversity, the construction of \textsc{CuraWeb} features three key techniques: domain-specific heuristic cleaning with customized thresholds, a two-stage soft semantic deduplication pipeline, and an automated Power Sampling mechanism that evaluates text structure and content value separately. Evaluated at a 1T-token scale, a 3B model trained on our dataset consistently outperforms models trained on strong baselines like FineWeb, Dolma3, and DCLM, especially on difficult reasoning tasks. Detailed ablation studies show that this joint optimization successfully achieves superior token efficiency, proving that maximizing the quality of each token yields better performance than training on unmanaged data. 
Overall, \textsc{CuraWeb} demonstrates that maximizing data quality while preserving diversity and broad coverage is much more effective than simply accumulating volume without precise selection.
We hope our dataset and methods can serve as a practical guide for future open-source data curation.

\newpage

%%%%%%%%%%%%%%%%%%%%%%%%%%%%%%%%%%%%%%%%%%%%%%%%%%%%%%%%%%%%

\bibliographystyle{unsrtnat}
\bibliography{references}

\begin{thebibliography}{48}
\providecommand{\natexlab}[1]{#1}
\providecommand{\url}[1]{\texttt{#1}}
\expandafter\ifx\csname urlstyle\endcsname\relax
  \providecommand{\doi}[1]{doi: #1}\else
  \providecommand{\doi}{doi: \begingroup \urlstyle{rm}\Url}\fi

\bibitem[Xi et~al.()Xi, Kong, Yang, Yang, Chen, Wang, Wang, Cai, Zhang, and Ye]{xi2025samplemix}
Xiangyu Xi, Deyang Kong, Jian Yang, Jiawei Yang, Zhengyu Chen, Wei Wang, Jingang Wang, Xunliang Cai, Shikun Zhang, and Wei Ye.
\newblock Samplemix: A sample-wise pre-training data mixing strategey by coordinating data quality and diversity.

\bibitem[Achiam et~al.(2023)Achiam, Adler, Agarwal, Ahmad, Akkaya, Aleman, Almeida, Altenschmidt, Altman, Anadkat, et~al.]{achiam2023gpt}
Josh Achiam, Steven Adler, Sandhini Agarwal, Lama Ahmad, Ilge Akkaya, Florencia~Leoni Aleman, Diogo Almeida, Janko Altenschmidt, Sam Altman, Shyamal Anadkat, et~al.
\newblock Gpt-4 technical report.
\newblock \emph{arXiv preprint arXiv:2303.08774}, 2023.

\bibitem[Liu et~al.(2024)Liu, Feng, Xue, Wang, Wu, Lu, Zhao, Deng, Zhang, Ruan, et~al.]{liu2024deepseek}
Aixin Liu, Bei Feng, Bing Xue, Bingxuan Wang, Bochao Wu, Chengda Lu, Chenggang Zhao, Chengqi Deng, Chenyu Zhang, Chong Ruan, et~al.
\newblock Deepseek-v3 technical report.
\newblock \emph{arXiv preprint arXiv:2412.19437}, 2024.

\bibitem[Yang et~al.(2025)Yang, Li, Yang, Zhang, Hui, Zheng, Yu, Gao, Huang, Lv, et~al.]{yang2025qwen3}
An~Yang, Anfeng Li, Baosong Yang, Beichen Zhang, Binyuan Hui, Bo~Zheng, Bowen Yu, Chang Gao, Chengen Huang, Chenxu Lv, et~al.
\newblock Qwen3 technical report.
\newblock \emph{arXiv preprint arXiv:2505.09388}, 2025.

\bibitem[Team et~al.(2025)Team, Li, Lei, Wang, Rong, Wang, Zhang, Gao, Zhang, Sun, et~al.]{team2025longcat}
Meituan~LongCat Team, Bei Li, Bingye Lei, Bo~Wang, Bolin Rong, Chao Wang, Chao Zhang, Chen Gao, Chen Zhang, Cheng Sun, et~al.
\newblock Longcat-flash technical report.
\newblock \emph{arXiv preprint arXiv:2509.01322}, 2025.

\bibitem[Grattafiori et~al.(2024)Grattafiori, Dubey, Jauhri, Pandey, Kadian, Al-Dahle, Letman, Mathur, Schelten, Vaughan, et~al.]{grattafiori2024llama}
Aaron Grattafiori, Abhimanyu Dubey, Abhinav Jauhri, Abhinav Pandey, Abhishek Kadian, Ahmad Al-Dahle, Aiesha Letman, Akhil Mathur, Alan Schelten, Alex Vaughan, et~al.
\newblock The llama 3 herd of models.
\newblock \emph{arXiv preprint arXiv:2407.21783}, 2024.

\bibitem[Adler et~al.(2024)Adler, Agarwal, Aithal, Anh, Bhattacharya, Brundyn, Casper, Catanzaro, Clay, Cohen, et~al.]{adler2024nemotron}
Bo~Adler, Niket Agarwal, Ashwath Aithal, Dong~H Anh, Pallab Bhattacharya, Annika Brundyn, Jared Casper, Bryan Catanzaro, Sharon Clay, Jonathan Cohen, et~al.
\newblock Nemotron-4 340b technical report.
\newblock \emph{arXiv preprint arXiv:2406.11704}, 2024.

\bibitem[Penedo et~al.(2024)Penedo, Kydl{\'\i}{\v{c}}ek, Lozhkov, Mitchell, Raffel, Von~Werra, Wolf, et~al.]{penedo2024fineweb}
Guilherme Penedo, Hynek Kydl{\'\i}{\v{c}}ek, Anton Lozhkov, Margaret Mitchell, Colin Raffel, Leandro Von~Werra, Thomas Wolf, et~al.
\newblock The fineweb datasets: Decanting the web for the finest text data at scale.
\newblock \emph{Advances in Neural Information Processing Systems}, 37:\penalty0 30811--30849, 2024.

\bibitem[Li et~al.(2024{\natexlab{a}})Li, Fang, Smyrnis, Ivgi, Jordan, Gadre, Bansal, Guha, Keh, Arora, et~al.]{li2024datacomp}
Jeffrey Li, Alex Fang, Georgios Smyrnis, Maor Ivgi, Matt Jordan, Samir Gadre, Hritik Bansal, Etash Guha, Sedrick Keh, Kushal Arora, et~al.
\newblock Datacomp-lm: In search of the next generation of training sets for language models.
\newblock \emph{Advances in Neural Information Processing Systems}, 37:\penalty0 14200--14282, 2024{\natexlab{a}}.

\bibitem[Rae et~al.(2021)Rae, Borgeaud, Cai, Millican, Hoffmann, Song, Aslanides, Henderson, Ring, Young, et~al.]{rae2021scaling}
Jack~W Rae, Sebastian Borgeaud, Trevor Cai, Katie Millican, Jordan Hoffmann, Francis Song, John Aslanides, Sarah Henderson, Roman Ring, Susannah Young, et~al.
\newblock Scaling language models: Methods, analysis \& insights from training gopher.
\newblock \emph{arXiv preprint arXiv:2112.11446}, 2021.

\bibitem[Penedo et~al.(2023)Penedo, Malartic, Hesslow, Cojocaru, Cappelli, Alobeidli, Pannier, Almazrouei, and Launay]{penedo2023refinedweb}
Guilherme Penedo, Quentin Malartic, Daniel Hesslow, Ruxandra Cojocaru, Alessandro Cappelli, Hamza Alobeidli, Baptiste Pannier, Ebtesam Almazrouei, and Julien Launay.
\newblock The refinedweb dataset for falcon llm: outperforming curated corpora with web data, and web data only.
\newblock \emph{arXiv preprint arXiv:2306.01116}, 2023.

\bibitem[Wenzek et~al.(2020)Wenzek, Lachaux, Conneau, Chaudhary, Guzm{\'a}n, Joulin, and Grave]{wenzek2020ccnet}
Guillaume Wenzek, Marie-Anne Lachaux, Alexis Conneau, Vishrav Chaudhary, Francisco Guzm{\'a}n, Armand Joulin, and Edouard Grave.
\newblock Ccnet: Extracting high quality monolingual datasets from web crawl data.
\newblock In \emph{Proceedings of the twelfth language resources and evaluation conference}, pages 4003--4012, 2020.

\bibitem[Huo et~al.(2025)Huo, Tu, Qin, Zheng, Zhang, Zhang, Li, Guo, Yao, Lou, et~al.]{huo2025dots}
Bi~Huo, Bin Tu, Cheng Qin, Da~Zheng, Debing Zhang, Dongjie Zhang, En~Li, Fu~Guo, Jian Yao, Jie Lou, et~al.
\newblock dots. llm1 technical report.
\newblock \emph{arXiv preprint arXiv:2506.05767}, 2025.

\bibitem[Cobbe et~al.(2021)Cobbe, Kosaraju, Bavarian, Chen, Jun, Kaiser, Plappert, Tworek, Hilton, Nakano, et~al.]{cobbe2021training}
Karl Cobbe, Vineet Kosaraju, Mohammad Bavarian, Mark Chen, Heewoo Jun, Lukasz Kaiser, Matthias Plappert, Jerry Tworek, Jacob Hilton, Reiichiro Nakano, et~al.
\newblock Training verifiers to solve math word problems.
\newblock \emph{arXiv preprint arXiv:2110.14168}, 2021.

\bibitem[Amini et~al.(2019)Amini, Gabriel, Lin, Koncel-Kedziorski, Choi, and Hajishirzi]{amini2019mathqa}
Aida Amini, Saadia Gabriel, Shanchuan Lin, Rik Koncel-Kedziorski, Yejin Choi, and Hannaneh Hajishirzi.
\newblock Mathqa: Towards interpretable math word problem solving with operation-based formalisms.
\newblock In \emph{Proceedings of the 2019 Conference of the North American Chapter of the Association for Computational Linguistics: Human Language Technologies, Volume 1 (Long and Short Papers)}, pages 2357--2367, 2019.

\bibitem[Hendrycks et~al.(2021)Hendrycks, Burns, Basart, Zou, Mazeika, Song, and Steinhardt]{hendrycks2021measuring}
Dan Hendrycks, Collin Burns, Steven Basart, Andy Zou, Mantas Mazeika, Dawn Song, and Jacob Steinhardt.
\newblock Measuring massive multitask language understanding.
\newblock In \emph{International Conference on Learning Representations}, 2021.
\newblock URL \url{https://openreview.net/forum?id=d7KBjmI3GmQ}.

\bibitem[Wang et~al.(2024)Wang, Ma, Zhang, Ni, Chandra, Guo, Ren, Arulraj, He, Jiang, et~al.]{wang2024mmlu}
Yubo Wang, Xueguang Ma, Ge~Zhang, Yuansheng Ni, Abhranil Chandra, Shiguang Guo, Weiming Ren, Aaran Arulraj, Xuan He, Ziyan Jiang, et~al.
\newblock Mmlu-pro: A more robust and challenging multi-task language understanding benchmark.
\newblock \emph{Advances in Neural Information Processing Systems}, 37:\penalty0 95266--95290, 2024.

\bibitem[Lai et~al.(2017)Lai, Xie, Liu, Yang, and Hovy]{lai2017race}
Guokun Lai, Qizhe Xie, Hanxiao Liu, Yiming Yang, and Eduard Hovy.
\newblock Race: Large-scale reading comprehension dataset from examinations.
\newblock In \emph{Proceedings of the 2017 conference on empirical methods in natural language processing}, pages 785--794, 2017.

\bibitem[Zellers et~al.(2019)Zellers, Holtzman, Bisk, Farhadi, and Choi]{zellers2019hellaswag}
Rowan Zellers, Ari Holtzman, Yonatan Bisk, Ali Farhadi, and Yejin Choi.
\newblock Hellaswag: Can a machine really finish your sentence?
\newblock In \emph{Proceedings of the 57th annual meeting of the association for computational linguistics}, pages 4791--4800, 2019.

\bibitem[Bisk et~al.(2020)Bisk, Zellers, Gao, Choi, et~al.]{bisk2020piqa}
Yonatan Bisk, Rowan Zellers, Jianfeng Gao, Yejin Choi, et~al.
\newblock Piqa: Reasoning about physical commonsense in natural language.
\newblock In \emph{Proceedings of the AAAI conference on artificial intelligence}, volume~34, pages 7432--7439, 2020.

\bibitem[Welbl et~al.(2017)Welbl, Liu, and Gardner]{welbl2017crowdsourcing}
Johannes Welbl, Nelson~F Liu, and Matt Gardner.
\newblock Crowdsourcing multiple choice science questions.
\newblock In \emph{Proceedings of the 3rd Workshop on Noisy User-generated Text}, pages 94--106, 2017.

\bibitem[Sakaguchi et~al.(2021)Sakaguchi, Bras, Bhagavatula, and Choi]{sakaguchi2021winogrande}
Keisuke Sakaguchi, Ronan~Le Bras, Chandra Bhagavatula, and Yejin Choi.
\newblock Winogrande: An adversarial winograd schema challenge at scale.
\newblock \emph{Communications of the ACM}, 64\penalty0 (9):\penalty0 99--106, 2021.

\bibitem[Mihaylov et~al.(2018)Mihaylov, Clark, Khot, and Sabharwal]{mihaylov2018can}
Todor Mihaylov, Peter Clark, Tushar Khot, and Ashish Sabharwal.
\newblock Can a suit of armor conduct electricity? a new dataset for open book question answering.
\newblock In \emph{Proceedings of the 2018 conference on empirical methods in natural language processing}, pages 2381--2391, 2018.

\bibitem[Olmo et~al.(2025)Olmo, Ettinger, Bertsch, Kuehl, Graham, Heineman, Groeneveld, Brahman, Timbers, Ivison, Morrison, Poznanski, Lo, Soldaini, Jordan, Chen, Noukhovitch, Lambert, Walsh, Dasigi, Berry, Malik, Shah, Geng, Arora, Gupta, Anderson, Xiao, Murray, Romero, Graf, Asai, Bhagia, Wettig, Liu, Rangapur, Anastasiades, Huang, Schwenk, Trivedi, Magnusson, Lochner, Liu, Miranda, Sap, Morgan, Schmitz, Guerquin, Wilson, Huff, Bras, Xin, Shao, Skjonsberg, Shen, Li, Wilde, Pyatkin, Merrill, Chang, Gu, Zeng, Sabharwal, Zettlemoyer, Koh, Farhadi, Smith, and Hajishirzi]{olmo2025olmo3}
Team Olmo, Allyson Ettinger, Amanda Bertsch, Bailey Kuehl, David Graham, David Heineman, Dirk Groeneveld, Faeze Brahman, Finbarr Timbers, Hamish Ivison, Jacob Morrison, Jake Poznanski, Kyle Lo, Luca Soldaini, Matt Jordan, Mayee Chen, Michael Noukhovitch, Nathan Lambert, Pete Walsh, Pradeep Dasigi, Robert Berry, Saumya Malik, Saurabh Shah, Scott Geng, Shane Arora, Shashank Gupta, Taira Anderson, Teng Xiao, Tyler Murray, Tyler Romero, Victoria Graf, Akari Asai, Akshita Bhagia, Alexander Wettig, Alisa Liu, Aman Rangapur, Chloe Anastasiades, Costa Huang, Dustin Schwenk, Harsh Trivedi, Ian Magnusson, Jaron Lochner, Jiacheng Liu, Lester James~V. Miranda, Maarten Sap, Malia Morgan, Michael Schmitz, Michal Guerquin, Michael Wilson, Regan Huff, Ronan~Le Bras, Rui Xin, Rulin Shao, Sam Skjonsberg, Shannon~Zejiang Shen, Shuyue~Stella Li, Tucker Wilde, Valentina Pyatkin, Will Merrill, Yapei Chang, Yuling Gu, Zhiyuan Zeng, Ashish Sabharwal, Luke Zettlemoyer, Pang~Wei Koh, Ali Farhadi, Noah~A. Smith, and Hannaneh
  Hajishirzi.
\newblock Olmo 3, 2025.
\newblock URL \url{https://arxiv.org/abs/2512.13961}.

\bibitem[Dodge et~al.(2021)Dodge, Sap, Marasovi{\'c}, Agnew, Ilharco, Groeneveld, Mitchell, and Gardner]{dodge2021documenting}
Jesse Dodge, Maarten Sap, Ana Marasovi{\'c}, William Agnew, Gabriel Ilharco, Dirk Groeneveld, Margaret Mitchell, and Matt Gardner.
\newblock Documenting large webtext corpora: A case study on the colossal clean crawled corpus.
\newblock In \emph{Proceedings of the 2021 conference on empirical methods in natural language processing}, pages 1286--1305, 2021.

\bibitem[Penedo et~al.(2025)Penedo, Kydl{\'\i}{\v{c}}ek, Sabol{\v{c}}ec, Messmer, Foroutan, Kargaran, Raffel, Jaggi, Werra, and Wolf]{penedo2025fineweb}
Guilherme Penedo, Hynek Kydl{\'\i}{\v{c}}ek, Vinko Sabol{\v{c}}ec, Bettina Messmer, Negar Foroutan, Amir~Hossein Kargaran, Colin Raffel, Martin Jaggi, Leandro~Von Werra, and Thomas Wolf.
\newblock Fineweb2: One pipeline to scale them all {\textemdash} adapting pre-training data processing to every language.
\newblock In \emph{Second Conference on Language Modeling}, 2025.
\newblock URL \url{https://openreview.net/forum?id=jnRBe6zatP}.

\bibitem[Broder(1997)]{broder1997resemblance}
Andrei~Z Broder.
\newblock On the resemblance and containment of documents.
\newblock In \emph{Proceedings. Compression and Complexity of SEQUENCES 1997 (Cat. No. 97TB100171)}, pages 21--29. IEEE, 1997.

\bibitem[Indyk and Motwani(1998)]{indyk1998approximate}
Piotr Indyk and Rajeev Motwani.
\newblock Approximate nearest neighbors: towards removing the curse of dimensionality.
\newblock In \emph{Proceedings of the thirtieth annual ACM symposium on Theory of computing}, pages 604--613, 1998.

\bibitem[Abbas et~al.(2023)Abbas, Tirumala, Simig, Ganguli, and Morcos]{abbas2023semdedup}
Amro Abbas, Kushal Tirumala, D{\'a}niel Simig, Surya Ganguli, and Ari~S Morcos.
\newblock Semdedup: Data-efficient learning at web-scale through semantic deduplication.
\newblock \emph{arXiv preprint arXiv:2303.09540}, 2023.

\bibitem[Lee et~al.(2022)Lee, Ippolito, Nystrom, Zhang, Eck, Callison-Burch, and Carlini]{lee2022deduplicating}
Katherine Lee, Daphne Ippolito, Andrew Nystrom, Chiyuan Zhang, Douglas Eck, Chris Callison-Burch, and Nicholas Carlini.
\newblock Deduplicating training data makes language models better.
\newblock In \emph{Proceedings of the 60th Annual Meeting of the Association for Computational Linguistics (Volume 1: Long Papers)}, pages 8424--8445, 2022.

\bibitem[Shen et~al.(2023)Shen, Tao, Ma, Neiswanger, Liu, Wang, Tan, Hestness, Vassilieva, Soboleva, et~al.]{shen2023slimpajama}
Zhiqiang Shen, Tianhua Tao, Liqun Ma, Willie Neiswanger, Zhengzhong Liu, Hongyi Wang, Bowen Tan, Joel Hestness, Natalia Vassilieva, Daria Soboleva, et~al.
\newblock Slimpajama-dc: Understanding data combinations for llm training.
\newblock \emph{arXiv preprint arXiv:2309.10818}, 2023.

\bibitem[Brown et~al.(2020)Brown, Mann, Ryder, Subbiah, Kaplan, Dhariwal, Neelakantan, Shyam, Sastry, Askell, et~al.]{brown2020language}
Tom Brown, Benjamin Mann, Nick Ryder, Melanie Subbiah, Jared~D Kaplan, Prafulla Dhariwal, Arvind Neelakantan, Pranav Shyam, Girish Sastry, Amanda Askell, et~al.
\newblock Language models are few-shot learners.
\newblock \emph{Advances in neural information processing systems}, 33:\penalty0 1877--1901, 2020.

\bibitem[Touvron et~al.(2023)Touvron, Lavril, Izacard, Martinet, Lachaux, Lacroix, Rozi{\`e}re, Goyal, Hambro, Azhar, et~al.]{touvron2023llama}
Hugo Touvron, Thibaut Lavril, Gautier Izacard, Xavier Martinet, Marie-Anne Lachaux, Timoth{\'e}e Lacroix, Baptiste Rozi{\`e}re, Naman Goyal, Eric Hambro, Faisal Azhar, et~al.
\newblock Llama: Open and efficient foundation language models.
\newblock \emph{arXiv preprint arXiv:2302.13971}, 2023.

\bibitem[Xie et~al.(2023)Xie, Pham, Dong, Du, Liu, Lu, Liang, Le, Ma, and Yu]{xie2023doremi}
Sang~Michael Xie, Hieu Pham, Xuanyi Dong, Nan Du, Hanxiao Liu, Yifeng Lu, Percy~S Liang, Quoc~V Le, Tengyu Ma, and Adams~Wei Yu.
\newblock Doremi: Optimizing data mixtures speeds up language model pretraining.
\newblock \emph{Advances in Neural Information Processing Systems}, 36:\penalty0 69798--69818, 2023.

\bibitem[Wettig et~al.(2025)Wettig, Lo, Min, Hajishirzi, Chen, and Soldaini]{wettig2025organize}
Alexander Wettig, Kyle Lo, Sewon Min, Hannaneh Hajishirzi, Danqi Chen, and Luca Soldaini.
\newblock Organize the web: Constructing domains enhances pre-training data curation.
\newblock \emph{arXiv preprint arXiv:2502.10341}, 2025.

\bibitem[Li et~al.(2024{\natexlab{b}})Li, Fang, Smyrnis, Ivgi, Jordan, Gadre, Bansal, Guha, Keh, Arora, Garg, Xin, Muennighoff, Heckel, Mercat, Chen, Gururangan, Wortsman, Albalak, Bitton, Nezhurina, Abbas, Hsieh, Ghosh, Gardner, Kilian, Zhang, Shao, Pratt, Sanyal, Ilharco, Daras, Marathe, Gokaslan, Zhang, Chandu, Nguyen, Vasiljevic, Kakade, Song, Sanghavi, Faghri, Oh, Zettlemoyer, Lo, El{-}Nouby, Pouransari, Toshev, Wang, Groeneveld, Soldaini, Koh, Jitsev, Kollar, Dimakis, Carmon, Dave, Schmidt, and Shankar]{DCLM-2024}
Jeffrey Li, Alex Fang, Georgios Smyrnis, Maor Ivgi, Matt Jordan, Samir~Yitzhak Gadre, Hritik Bansal, Etash~Kumar Guha, Sedrick~Scott Keh, Kushal Arora, Saurabh Garg, Rui Xin, Niklas Muennighoff, Reinhard Heckel, Jean Mercat, Mayee~F. Chen, Suchin Gururangan, Mitchell Wortsman, Alon Albalak, Yonatan Bitton, Marianna Nezhurina, Amro Abbas, Cheng{-}Yu Hsieh, Dhruba Ghosh, Josh Gardner, Maciej Kilian, Hanlin Zhang, Rulin Shao, Sarah~M. Pratt, Sunny Sanyal, Gabriel Ilharco, Giannis Daras, Kalyani Marathe, Aaron Gokaslan, Jieyu Zhang, Khyathi~Raghavi Chandu, Thao Nguyen, Igor Vasiljevic, Sham~M. Kakade, Shuran Song, Sujay Sanghavi, Fartash Faghri, Sewoong Oh, Luke Zettlemoyer, Kyle Lo, Alaaeldin El{-}Nouby, Hadi Pouransari, Alexander Toshev, Stephanie Wang, Dirk Groeneveld, Luca Soldaini, Pang~Wei Koh, Jenia Jitsev, Thomas Kollar, Alex Dimakis, Yair Carmon, Achal Dave, Ludwig Schmidt, and Vaishaal Shankar.
\newblock Datacomp-lm: In search of the next generation of training sets for language models.
\newblock In Amir Globersons, Lester Mackey, Danielle Belgrave, Angela Fan, Ulrich Paquet, Jakub~M. Tomczak, and Cheng Zhang, editors, \emph{Advances in Neural Information Processing Systems 38: Annual Conference on Neural Information Processing Systems 2024, NeurIPS 2024, Vancouver, BC, Canada, December 10 - 15, 2024}, 2024{\natexlab{b}}.
\newblock URL \url{http://papers.nips.cc/paper\_files/paper/2024/hash/19e4ea30dded58259665db375885e412-Abstract-Datasets\_and\_Benchmarks\_Track.html}.

\bibitem[Bevendorff et~al.(2018)Bevendorff, Stein, Hagen, and Potthast]{bevendorff2018elastic}
Janek Bevendorff, Benno Stein, Matthias Hagen, and Martin Potthast.
\newblock Elastic chatnoir: Search engine for the clueweb and the common crawl.
\newblock In \emph{European conference on information retrieval}, pages 820--824. Springer, 2018.

\bibitem[Joulin et~al.(2017)Joulin, Grave, Bojanowski, and Mikolov]{joulin2017bag}
Armand Joulin, Edouard Grave, Piotr Bojanowski, and Tom{\'a}{\v{s}} Mikolov.
\newblock Bag of tricks for efficient text classification.
\newblock In \emph{Proceedings of the 15th conference of the European chapter of the association for computational linguistics: volume 2, short papers}, pages 427--431, 2017.

\bibitem[Longpre et~al.(2024)Longpre, Yauney, Reif, Lee, Roberts, Zoph, Zhou, Wei, Robinson, Mimno, et~al.]{longpre2024pretrainer}
Shayne Longpre, Gregory Yauney, Emily Reif, Katherine Lee, Adam Roberts, Barret Zoph, Denny Zhou, Jason Wei, Kevin Robinson, David Mimno, et~al.
\newblock A pretrainer’s guide to training data: Measuring the effects of data age, domain coverage, quality, \& toxicity.
\newblock In \emph{Proceedings of the 2024 Conference of the North American Chapter of the Association for Computational Linguistics: Human Language Technologies (Volume 1: Long Papers)}, pages 3245--3276, 2024.

\bibitem[Hurst et~al.(2024)Hurst, Lerer, Goucher, Perelman, Ramesh, Clark, Ostrow, Welihinda, Hayes, Radford, et~al.]{hurst2024gpt}
Aaron Hurst, Adam Lerer, Adam~P Goucher, Adam Perelman, Aditya Ramesh, Aidan Clark, AJ~Ostrow, Akila Welihinda, Alan Hayes, Alec Radford, et~al.
\newblock Gpt-4o system card.
\newblock \emph{arXiv preprint arXiv:2410.21276}, 2024.

\bibitem[Kendall et~al.(2018)Kendall, Gal, and Cipolla]{kendall2018multi}
Alex Kendall, Yarin Gal, and Roberto Cipolla.
\newblock Multi-task learning using uncertainty to weigh losses for scene geometry and semantics.
\newblock In \emph{2018 {IEEE} Conference on Computer Vision and Pattern Recognition, {CVPR} 2018, Salt Lake City, UT, USA, June 18-22, 2018}, pages 7482--7491. Computer Vision Foundation / {IEEE} Computer Society, 2018.
\newblock \doi{10.1109/CVPR.2018.00781}.
\newblock URL \url{http://openaccess.thecvf.com/content\_cvpr\_2018/html/Kendall\_Multi-Task\_Learning\_Using\_CVPR\_2018\_paper.html}.

\bibitem[Ethayarajh(2019)]{ethayarajh2019contextual}
Kawin Ethayarajh.
\newblock How contextual are contextualized word representations? comparing the geometry of bert, elmo, and gpt-2 embeddings.
\newblock In \emph{Proceedings of the 2019 conference on empirical methods in natural language processing and the 9th international joint conference on natural language processing (EMNLP-IJCNLP)}, pages 55--65, 2019.

\bibitem[Sturua et~al.(2024)Sturua, Mohr, Akram, G{\"u}nther, Wang, Krimmel, Wang, Mastrapas, Koukounas, Wang, et~al.]{sturua2024jina}
Saba Sturua, Isabelle Mohr, Mohammad~Kalim Akram, Michael G{\"u}nther, Bo~Wang, Markus Krimmel, Feng Wang, Georgios Mastrapas, Andreas Koukounas, Nan Wang, et~al.
\newblock jina-embeddings-v3: Multilingual embeddings with task lora.
\newblock \emph{arXiv preprint arXiv:2409.10173}, 2024.

\bibitem[Team et~al.(2024)Team, Mesnard, Hardin, Dadashi, Bhupatiraju, Pathak, Sifre, Rivi{\`e}re, Kale, Love, et~al.]{team2024gemma}
Gemma Team, Thomas Mesnard, Cassidy Hardin, Robert Dadashi, Surya Bhupatiraju, Shreya Pathak, Laurent Sifre, Morgane Rivi{\`e}re, Mihir~Sanjay Kale, Juliette Love, et~al.
\newblock Gemma: Open models based on gemini research and technology.
\newblock \emph{arXiv preprint arXiv:2403.08295}, 2024.

\bibitem[Su et~al.(2025)Su, Kong, Lin, Jennings, Norick, Kliegl, Patwary, Shoeybi, and Catanzaro]{Nemotron-CC-2025}
Dan Su, Kezhi Kong, Ying Lin, Joseph Jennings, Brandon Norick, Markus Kliegl, Mostofa Patwary, Mohammad Shoeybi, and Bryan Catanzaro.
\newblock Nemotron-cc: Transforming common crawl into a refined long-horizon pretraining dataset.
\newblock In Wanxiang Che, Joyce Nabende, Ekaterina Shutova, and Mohammad~Taher Pilehvar, editors, \emph{Proceedings of the 63rd Annual Meeting of the Association for Computational Linguistics (Volume 1: Long Papers), {ACL} 2025, Vienna, Austria, July 27 - August 1, 2025}, pages 2459--2475. Association for Computational Linguistics, 2025.
\newblock URL \url{https://aclanthology.org/2025.acl-long.123/}.

\bibitem[Soldaini et~al.(2024)Soldaini, Kinney, Bhagia, Schwenk, Atkinson, Authur, Bogin, Chandu, Dumas, Elazar, et~al.]{soldaini2024dolma}
Luca Soldaini, Rodney Kinney, Akshita Bhagia, Dustin Schwenk, David Atkinson, Russell Authur, Ben Bogin, Khyathi Chandu, Jennifer Dumas, Yanai Elazar, et~al.
\newblock Dolma: An open corpus of three trillion tokens for language model pretraining research.
\newblock In \emph{Proceedings of the 62nd Annual Meeting of the Association for Computational Linguistics (Volume 1: Long Papers)}, pages 15725--15788, 2024.

\bibitem[Gao et~al.(2023)Gao, Tow, Abbasi, Biderman, Black, DiPofi, Foster, Golding, Hsu, Le~Noac'h, Li, McDonell, Muennighoff, Ociepa, Phang, Reynolds, Schoelkopf, Skowron, Sutawika, Tang, Thite, Wang, Wang, and Zou]{eval-harness}
Leo Gao, Jonathan Tow, Baber Abbasi, Stella Biderman, Sid Black, Anthony DiPofi, Charles Foster, Laurence Golding, Jeffrey Hsu, Alain Le~Noac'h, Haonan Li, Kyle McDonell, Niklas Muennighoff, Chris Ociepa, Jason Phang, Laria Reynolds, Hailey Schoelkopf, Aviya Skowron, Lintang Sutawika, Eric Tang, Anish Thite, Ben Wang, Kevin Wang, and Andy Zou.
\newblock A framework for few-shot language model evaluation, 12 2023.
\newblock URL \url{https://zenodo.org/records/10256836}.

\bibitem[Kingma and Ba(2014)]{kingma2014adam}
Diederik~P Kingma and Jimmy Ba.
\newblock Adam: A method for stochastic optimization.
\newblock \emph{arXiv preprint arXiv:1412.6980}, 2014.

\end{thebibliography}

\newpage

\appendix

\subsection*{A \quad Training Configuration}
Table~\ref{tab:training_config} lists the architecture and optimization hyperparameters used for training. 
\begin{table}[h]
\centering
\caption{Training hyperparameters.}
\label{tab:training_config}
\begin{tabular}{ll}
\toprule
\textbf{Hyperparameter} & \textbf{Value} \\
\midrule
Parameters             & 3B \\
Architecture           & LLaMA (RMSNorm, SwiGLU, RoPE, GQA) \\
Layers                 & 32 \\
Hidden size            & 4096 \\
FFN hidden size        & 14336 \\
Attention heads        & 32 \\
KV heads (GQA)         & 8 \\
Sequence length        & 8192 \\
Vocabulary size        & 131072 \\
Precision              & BF16 \\
Optimizer              & Adam \cite{kingma2014adam}($\beta_1{=}0.9$, $\beta_2{=}0.95$, $\epsilon{=}10^{-8}$) \\
Learning rate          & $3 \times 10^{-4}$ (cosine decay, min $3 \times 10^{-5}$) \\
Warmup steps           & 122,070 \\
Global batch size      & 960 \\
Weight decay           & 0.1 \\
Gradient clipping      & 1.0 \\
Training tokens        & 200B / 1T \\
\bottomrule
\end{tabular}
\end{table}

\subsection*{B \quad Domain Taxonomy}

Table~\ref{tab:domain_l1} presents the 26 primary domain (L1-domain) categories utilized in our data understanding system. 
Adapted from the Google AdSense classification hierarchy, each L1 category is further subdivided into fine-grained secondary (L2) subcategories, yielding a total of 105 distinct categories.

\begin{table}[h]
\centering
\caption{26 L1-domain categories.}
\label{tab:domain_l1}
\begin{tabular}{clcl}
\toprule
\textbf{ID} & \textbf{Domain} & \textbf{ID} & \textbf{Domain} \\
\midrule
1  & \texttt{Adult}                     & 14 & \texttt{Internet\_and\_Telecom} \\
2  & \texttt{Arts\_and\_Entertainment}   & 15 & \texttt{Jobs\_and\_Education} \\
3  & \texttt{Autos\_and\_Vehicles}      & 16 & \texttt{Law\_and\_Government} \\
4  & \texttt{Beauty\_and\_Fitness}      & 17 & \texttt{News} \\
5  & \texttt{Books\_and\_Literature}    & 18 & \texttt{Online\_Communities} \\
6  & \texttt{Business\_and\_Industrial} & 19 & \texttt{People\_and\_Society} \\
7  & \texttt{Computers\_and\_Electronics} & 20 & \texttt{Pets\_and\_Animals} \\
8  & \texttt{Finance}                   & 21 & \texttt{Real\_Estate} \\
9  & \texttt{Food\_and\_Drink}          & 22 & \texttt{Science} \\
10 & \texttt{Games}                     & 23 & \texttt{Sensitive\_Subjects} \\
11 & \texttt{Health}                    & 24 & \texttt{Shopping} \\
12 & \texttt{Hobbies\_and\_Leisure}       & 25 & \texttt{Sports} \\
13 & \texttt{Home\_and\_Garden}         & 26 & \texttt{Travel\_and\_Transportation} \\
\bottomrule
\end{tabular}
\end{table}

\subsection*{C \quad Annotation Prompts}
    \label{app:annotation_prompts}
    
    \paragraph{Quality Scoring Prompt}
    \label{app:quality_scoring_prompt}
    
    The following prompt is used to obtain supervision labels for the quality dimensions. Table~\ref{tab:quality_rubric} summarizes the scoring rubric, where all dimensions use a 0--3 scale.
    
    \begin{promptbox}{Quality Scoring Prompt}
    You are a large language model training data annotator tasked with evaluating the quality of text data and scoring the text from each of the following dimensions separately.
    
    The dimensions include four expression-related dimensions: Writing, Coherence, Completeness, and Logicality; four value-oriented dimensions: Knowledge, EducationValue, PracticalHelpfulness, and Reasoning; and one safety-related dimension: Safety.
    
    Name: Writing
    Description: Evaluate the spelling, grammar, punctuation, and overall writing quality of the text.
    Scoring Criteria:
    0: Severe Writing Errors: the text is either unreadable or incomprehensible because the content is dominated by serious writing errors, including garbled text, abnormal symbols, lack of proper spacing/punctuation, repetitive content, or grammar/spelling mistakes.
    1: Readable with Obvious Errors: the text contains multiple obvious errors in spelling, grammar, or punctuation that interrupt the flow of reading and impede overall comprehension, but the content remains generally readable and understandable.
    2: Good Writing with Minor Flaws: the text is well-written with only a few ignorable mistakes that have little impact on reading and understanding. The writing flows naturally and effectively communicates its message despite these small imperfections.
    3: Excellent Writing: the text demonstrates exemplary writing quality without writing errors. It is polished, precise, and highly effective in its use of language.

    ...

    Requirements:
    Evaluate the texts based on the above dimensions, provide evaluation reasoning first, and then give a quality evaluation score. Output format in JSON:
    
    {
      "Evaluation Reasoning": 
      {
        "Writing": "...",
        "Coherence": "...",
        "Completeness": "...",
        "Logicality": "...",
        "Knowledge": "...",
        "EducationValue": "...",
        "PracticalHelpfulness": "...",
        "Reasoning": "...",
        "Safety": "..."
      },
      "Writing": X,
      "Coherence": X,
      "Completeness": X,
      "Logicality": X,
      "Knowledge": X,
      "EducationValue": X,
      "PracticalHelpfulness": X,
      "Reasoning": X,
      "Safety": X
    }

    Evaluate the following text as a whole: $TEXT$
    \end{promptbox}

    \begin{longtable}{P{0.16\textwidth}P{0.78\textwidth}}
    \caption{Scoring rubric for quality annotation dimensions.}
    \label{tab:quality_rubric}\\
    
    \toprule
    \textbf{Dimension} & \textbf{Scoring Criteria} \\
    \midrule
    \endfirsthead
    
    \toprule
    \textbf{Dimension} & \textbf{Scoring Criteria} \\
    \midrule
    \endhead
    
    \midrule
    \endfoot
    
    \bottomrule
    \endlastfoot
    
    Writing &
    \scoreitem{0}{Severe writing errors; the text is unreadable or incomprehensible due to garbled text, abnormal symbols, repetition, or serious grammar/spelling errors.}
    \scoreitem{1}{Readable with obvious errors; multiple spelling, grammar, or punctuation errors interrupt reading and impede comprehension.}
    \scoreitem{2}{Good writing with minor flaws; only a few negligible mistakes that do not affect understanding.}
    \scoreitem{3}{Excellent writing; polished, precise, and free of writing errors.}
    \\
    \midrule
    Coherence &
    \scoreitem{0}{Severely incoherent; ideas are randomly arranged with abrupt shifts between unrelated or conflicting topics.}
    \scoreitem{1}{Partially coherent; some basic connections exist, but poor transitions and fragmented organization make the topic hard to follow.}
    \scoreitem{2}{Generally coherent; ideas are mostly logically ordered with only minor disruptions or inconsistencies.}
    \scoreitem{3}{Excellent coherence; ideas flow smoothly with clear thematic development and effective transitions.}
    \\
    \midrule
    Completeness &
    \scoreitem{0}{Severely incomplete; major necessary components are missing, leaving substantial unanswered questions.}
    \scoreitem{1}{Partially complete; basic information is present, but core messages or important components are omitted or underdeveloped.}
    \scoreitem{2}{Generally complete; most key points are covered, with only minor missing details.}
    \scoreitem{3}{Comprehensively complete; all necessary points are addressed with appropriate depth and cohesion.}
    \\
    \midrule
    Logicality &
    \scoreitem{0}{Severe logic errors; the text contains contradictions, unclear causality, fallacies, or unsupported claims.}
    \scoreitem{1}{Partially logical; basic consistency exists, but flaws, insufficient support, or contradictions weaken the argument.}
    \scoreitem{2}{Generally sound logic; arguments are mostly valid and supported, with only minor lapses.}
    \scoreitem{3}{Exemplary logical reasoning; claims are well-supported, connections are explicit, and no contradictions or fallacies are present.}
    \\
    \midrule
    Knowledge &
    \scoreitem{0}{Minimal knowledge value; the text is superficial, sparse, or mostly uninformative.}
    \scoreitem{1}{Basic knowledge with limited depth; factually adequate but lacks analytical depth, novelty, or conceptual integration.}
    \scoreitem{2}{Substantial knowledge; demonstrates good conceptual understanding, professional terminology, and meaningful connections.}
    \scoreitem{3}{Advanced knowledge; shows exceptional depth, originality, professional insight, and sophisticated conceptual integration.}
    \\
    \midrule
    EducationValue &
    \scoreitem{0}{Negligible educational value; irrelevant to education or dominated by inappropriate or promotional content.}
    \scoreitem{1}{Limited educational utility; contains some relevant information but is poorly aligned, disorganized, or unsuitable for teaching.}
    \scoreitem{2}{Moderately educational; suitable for school-level use, with clear presentation and substantial educational content.}
    \scoreitem{3}{Excellent educational resource; well-suited for teaching, with clear reasoning, learning value, and appropriate depth.}
    \\
    \midrule
    Practical &
    \scoreitem{0}{No practical value; vague, irrelevant, or lacking actionable guidance and implementable suggestions.}
    \scoreitem{1}{Limited practical application; contains some useful elements but lacks specificity, guidance, or adaptability.}
    \scoreitem{2}{Moderately helpful; provides actionable steps, useful recommendations, and problem-solving structure.}
    \scoreitem{3}{Exceptionally practical; offers clear step-by-step instructions, concrete examples, and executable methods.}
    \\
    \midrule
    Reasoning &
    \scoreitem{0}{No reasoning value; contains no reasoning-related content or has fundamental reasoning errors.}
    \scoreitem{1}{Limited reasoning value; attempts reasoning but includes noticeable errors or conclusions unsupported by premises.}
    \scoreitem{2}{Generally sound reasoning; mostly accurate reasoning with appropriate evidence-conclusion connections.}
    \scoreitem{3}{Exemplary reasoning; rigorous, methodical, well-supported, and free from logical fallacies.}
    \\
    \midrule
    Safety &
    \scoreitem{0}{Unsafe content; involves serious issues such as cheating, drugs, gambling, pornography, violence, horror, anti-human activity, or illegal transactions.}
    \scoreitem{1}{Mainly safe; the main content is unrelated to unsafe issues, even if a few unsafe phrases appear.}
    \\
    
    \end{longtable}

  \paragraph*{Domain Classification Prompt}
    \label{app:domain_classification_prompt}
    
    The following prompt is used to classify web page contents into predefined
    domain categories.
    
    \begin{promptbox}{Domain Classification Prompt}
    You are a highly skilled web page domain classifier. Your task is to identify the correct domain for a given web page content based on predefined categories.
    
    The top-level domains are:
      1. Adult
      2. Arts_and_Entertainment
      3. Autos_and_Vehicles
      4. Beauty_and_Fitness
      5. Books_and_Literature
      6. Business_and_Industrial
      7. Computers_and_Electronics
      8. Finance
      9. Food_and_Drink
      10. Games
      11. Health
      12. Hobbies_and_Leisure
      13. Home_and_Garden
      14. Internet_and_Telecom
      15. Jobs_and_Education
      16. Law_and_Government
      17. News
      18. Online_Communities
      19. People_and_Society
      20. Pets_and_Animals
      21. Real_Estate
      22. Science
      23. Sensitive_Subjects
      24. Shopping
      25. Sports
      26. Travel_and_Transportation
      
    The second-level domains are:
    ...
    
    Each domain is structured as:
    /PARENT_DOMAIN/SUB_DOMAIN
    \end{promptbox}

\subsection*{D \quad Comprehensive Comparison of Gopher and CuraWeb Heuristic Rules}
Table~\ref{tab:full_rules} presents the complete set of heuristic filtering rules, comparing the original Gopher~\cite{rae2021scaling} thresholds with our recalibrated thresholds. 
Rules designated as ``Removed'' were eliminated due to an excessive risk of over-filtering high-quality content, while those labeled ``Retained'' maintain their original thresholds.
 
\begin{table}[h]
\centering
\caption{Complete heuristic rule comparison: Gopher original vs.\ CuraWeb optimized.}
\label{tab:full_rules}
\resizebox{\textwidth}{!}{%
\begin{tabular}{llllp{5cm}}
\toprule
\textbf{Rule} & \textbf{Gopher Original} & \textbf{CuraWeb} & \textbf{Status} & \textbf{Motivation} \\
\midrule
\multicolumn{5}{l}{\textit{Document-level rules}} \\
word\_count (min)             & $\geq 50$    & $\geq 25$   & Relaxed  & Retain concise knowledge snippets \\
word\_count (max)             & $\leq 100{,}000$ & $\leq 100{,}000$ & Retained & --- \\
mean\_word\_length            & $3\text{--}10$    & $3\text{--}10$   & Retained & --- \\ 
symbol\_to\_word\_ratio (hash/ellipsis) & $> 0.1$ & $> 0.38$ & Relaxed & Tolerate symbol-dense formulaic content \\
alphabetic\_char\_ratio       & $< 0.80$     & $< 0.26$    & Relaxed  & Prevent STEM data loss from LaTeX/formulas \\
stop\_word\_filter            & $\geq 2$ stop words & $\geq 2$ stop words & Retained & --- \\
\midrule
\multicolumn{5}{l}{\textit{Removed rules (high over-filtering risk)}} \\
median\_word\_length          & $\leq 3$     & ---         & Removed  & Reduces misjudgment of short-word technical texts \\
required\_word\_count         & $\leq 2$     & ---         & Removed  & Eliminates redundant low-precision constraint \\
transformed\_text\_len        & $\geq 10{,}000$ & ---      & Removed  & Prevents purging of long technical specifications \\
fraction\_bullet\_points      & $\geq 0.9$   & ---         & Removed  & Preserves well-structured technical outlines \\
fraction\_ellipses            & $\geq 0.3$   & ---         & Removed  & Removes low-precision constraint \\ 
\midrule
\multicolumn{5}{l}{\textit{Repetition removal rules (retained from Gopher, Table A1)}} \\
Duplicate line fraction      & $> 0.30$     & $> 0.30$    & Retained & --- \\
Duplicate paragraph fraction & $> 0.30$     & $> 0.30$    & Retained & --- \\
Duplicate line char fraction & $> 0.20$     & $> 0.20$    & Retained & --- \\
Duplicate paragraph char fraction & $> 0.20$ & $> 0.20$  & Retained & --- \\
Top 2-gram char fraction     & $> 0.20$     & $> 0.20$    & Retained & --- \\
Top 3-gram char fraction     & $> 0.18$     & $> 0.18$    & Retained & --- \\
Top 4-gram char fraction     & $> 0.16$     & $> 0.16$    & Retained & --- \\
Duplicate 5--10-gram char fraction & $> 0.15\text{--}0.10$ & $> 0.15\text{--}0.10$ & Retained & --- \\
\midrule
\multicolumn{5}{l}{\textit{STEM bypass mechanism (CuraWeb only)}} \\
Mathematics / STEM / Code domains & N/A & Skip rule filter & Added & Directly forward to model-based filter \\
\bottomrule 
\end{tabular}%
}
\end{table}

% 请确保你的 LaTex 导言区（Preamble）包含以下宏包：
% \usepackage{booktabs}
% \usepackage{algorithm}
% \usepackage{algpseudocode}
% \usepackage{amsmath}
% \usepackage{amssymb}

\subsection*{E \quad Soft Deduplication Algorithm}
\label{subsec:Soft Deduplication Algorithm}

To strictly bound joint false positive propagation inherent in single-edge hard thresholds, we propose a soft deduplication mechanism. This approach introduces a multi-path voting scheme where a document is classified as a duplicate only when its accumulated evidence across multiple edges exceeds a specific target threshold, rather than relying on a single high-similarity match.

Algorithm~\ref{alg:soft_dedup} details the complete soft Deduplication procedure. Operating cluster-by-cluster on the output of K-Means semantic clustering, the algorithm processes documents in batches. For each current batch of candidates $C$, intra-batch conflicts are first resolved via a strict hard threshold $t_{\text{strict}}$, retaining earlier indices as valid leaders $\mathcal{L}$. 

Each valid leader $\ell \in \mathcal{L}$ then updates the penalty scores of all remaining unprocessed target nodes $u \in U$. The accumulated penalty score $S[u]$ is incremented by a dynamic interval weight $w^*$, determined by the maximum similarity threshold $t_k$ satisfied by the edge:
\begin{equation}
w^* = \max\bigl\{w_k \mid \mathrm{sim}(E_\ell, E_u) \geq t_k,\ (t_k,w_k)\in\mathcal{T}\bigr\}
\end{equation}
A node is marked as a duplicate ($\texttt{is\_duplicate}[u] \gets 1$) if and only if $S[u] \geq T_{\text{target}}$ (set to $10$ in our experiments). Because the probability of accumulating $T_{\text{target}}$ points from independent false-positive edges decays exponentially with the number of required confirmations, this multi-path voting mechanism effectively prevents error propagation.

\begin{algorithm}[t]
\caption{The Soft Deduplication Algorithm via Weighted Score Accumulation}
\label{alg:soft_dedup}
\begin{algorithmic}[1]
\Require Embedding matrix $E \in \mathbb{R}^{N \times d}$ (L2-normalized),
         similarity interval weights $\mathcal{T} = \{(t_1, w_1), \dots, (t_K, w_K)\}$,
         target score $T_{\text{target}}$,
         strict threshold $t_{\text{strict}}$,
         batch size $B$
\Ensure Binary array $\texttt{is\_duplicate} \in \{0,1\}^N$

\State $\texttt{is\_processed} \gets \mathbf{0}^N$
\State $\texttt{is\_duplicate} \gets \mathbf{0}^N$
\State $S \gets \mathbf{0}^N$ \Comment{Accumulated penalty scores}

\While{$\exists\, i : \texttt{is\_processed}[i] = 0$}
    \State $U \gets \{i \mid \texttt{is\_processed}[i] = 0\}$ \Comment{Unprocessed nodes}
    \State $C \gets U[\,{:}\,B\,]$ \Comment{Current batch of candidates}
    
    \State \textbf{// Step 1: Intra-batch conflict resolution}
    \For{each pair $i < j$ with $i, j \in C$}
        \If{$\mathrm{sim}(E_i,\, E_j) \geq t_{\text{strict}}$}
            \State kill $j$ within batch \Comment{Retain earlier index as leader}
        \EndIf
    \EndFor
    \State $\mathcal{L} \gets \{i \in C \mid i \text{ not killed}\}$ \Comment{Valid leaders}
    
    \State \textbf{// Step 2: Score accumulation}
    \For{each leader $\ell \in \mathcal{L}$}
        \For{each unprocessed target $u \in U,\ u \neq \ell$}
            \State $w^* \gets \max\bigl\{w_k \mid \mathrm{sim}(E_\ell, E_u) \geq t_k,\ (t_k,w_k)\in\mathcal{T}\bigr\}$
            \State $S[u] \mathrel{+}= w^*$
            \If{$S[u] \geq T_{\text{target}}$}
                \State $\texttt{is\_duplicate}[u] \gets 1$
            \EndIf
        \EndFor
    \EndFor
    
    \State $\texttt{is\_processed}\bigl[C \cup \{u \mid \texttt{is\_duplicate}[u]=1\}\bigr] \gets 1$
\EndWhile

\State \Return $\texttt{is\_duplicate}$
\end{algorithmic}
\end{algorithm}

\begin{figure}[h]
  \centering
  
  \begin{center}
    \begin{minipage}[t]{0.31\textwidth}
      \centering
      \includegraphics[width=\linewidth]{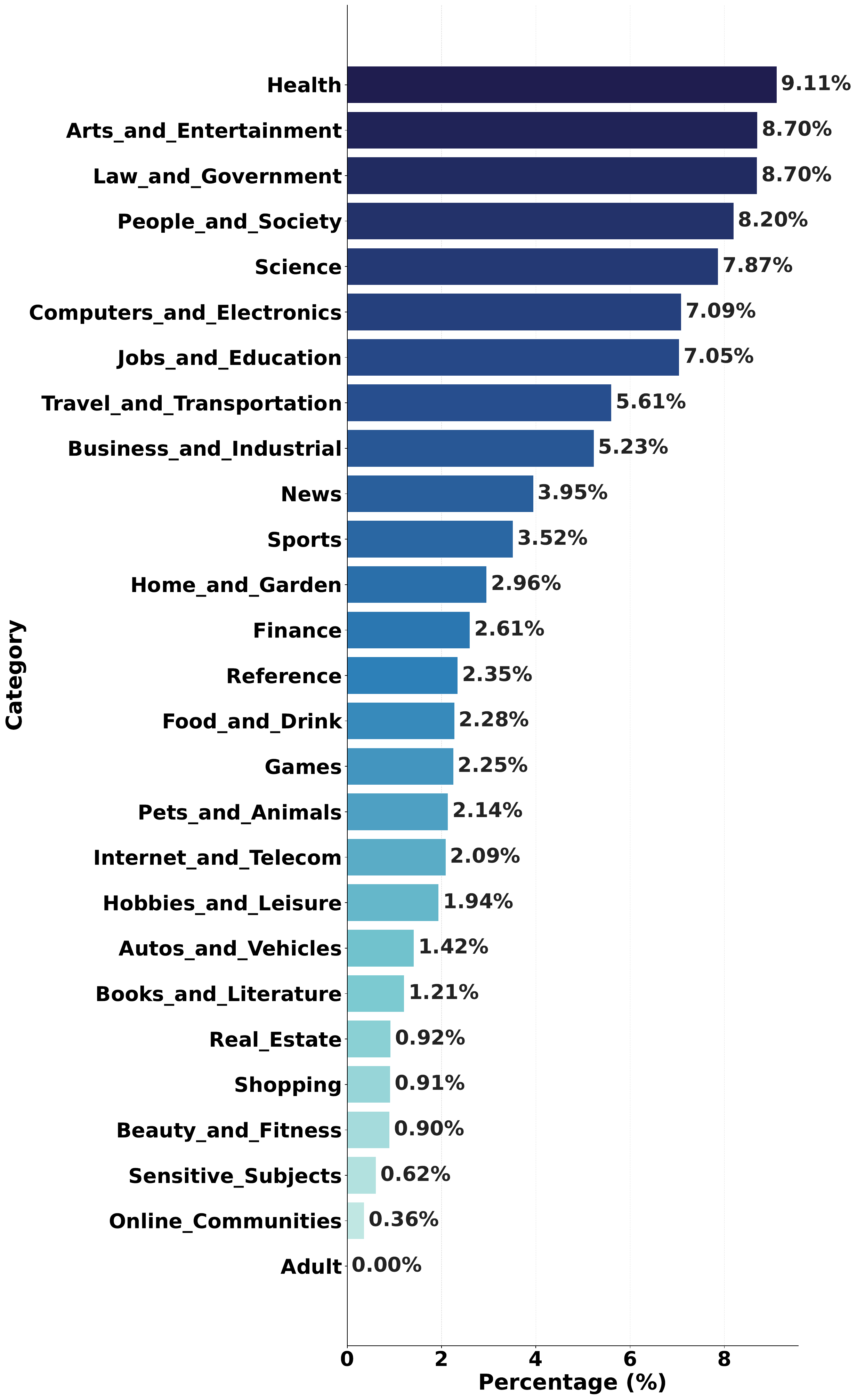}\\[-0.2em]
      {\small CuraWeb}
    \end{minipage}
    \hfill
    \begin{minipage}[t]{0.31\textwidth}
      \centering
      \includegraphics[width=\linewidth]{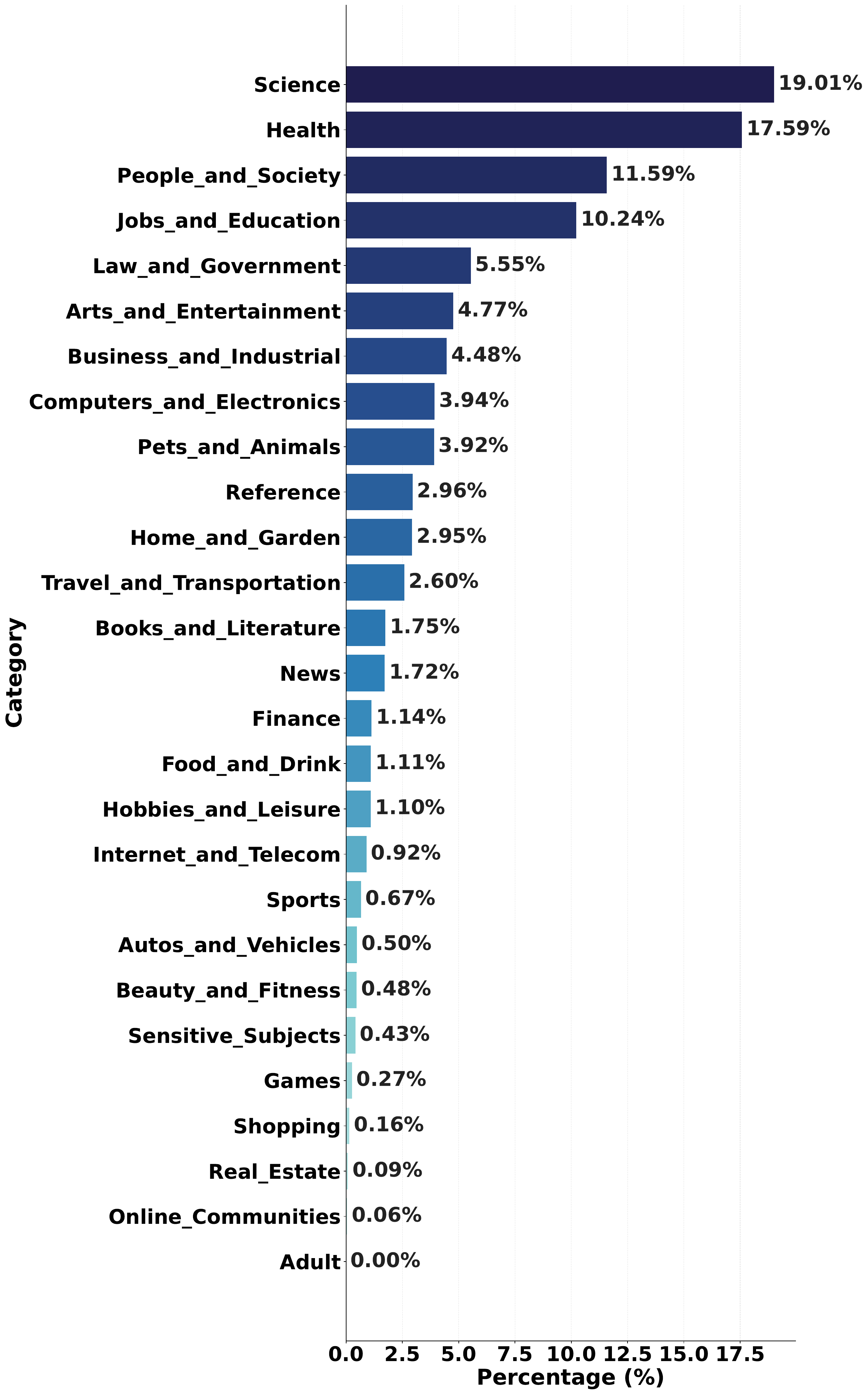}\\[-0.2em]
      {\small FineWeb-Edu}
    \end{minipage}
    \hfill
    \begin{minipage}[t]{0.31\textwidth}
      \centering
      \includegraphics[width=\linewidth]{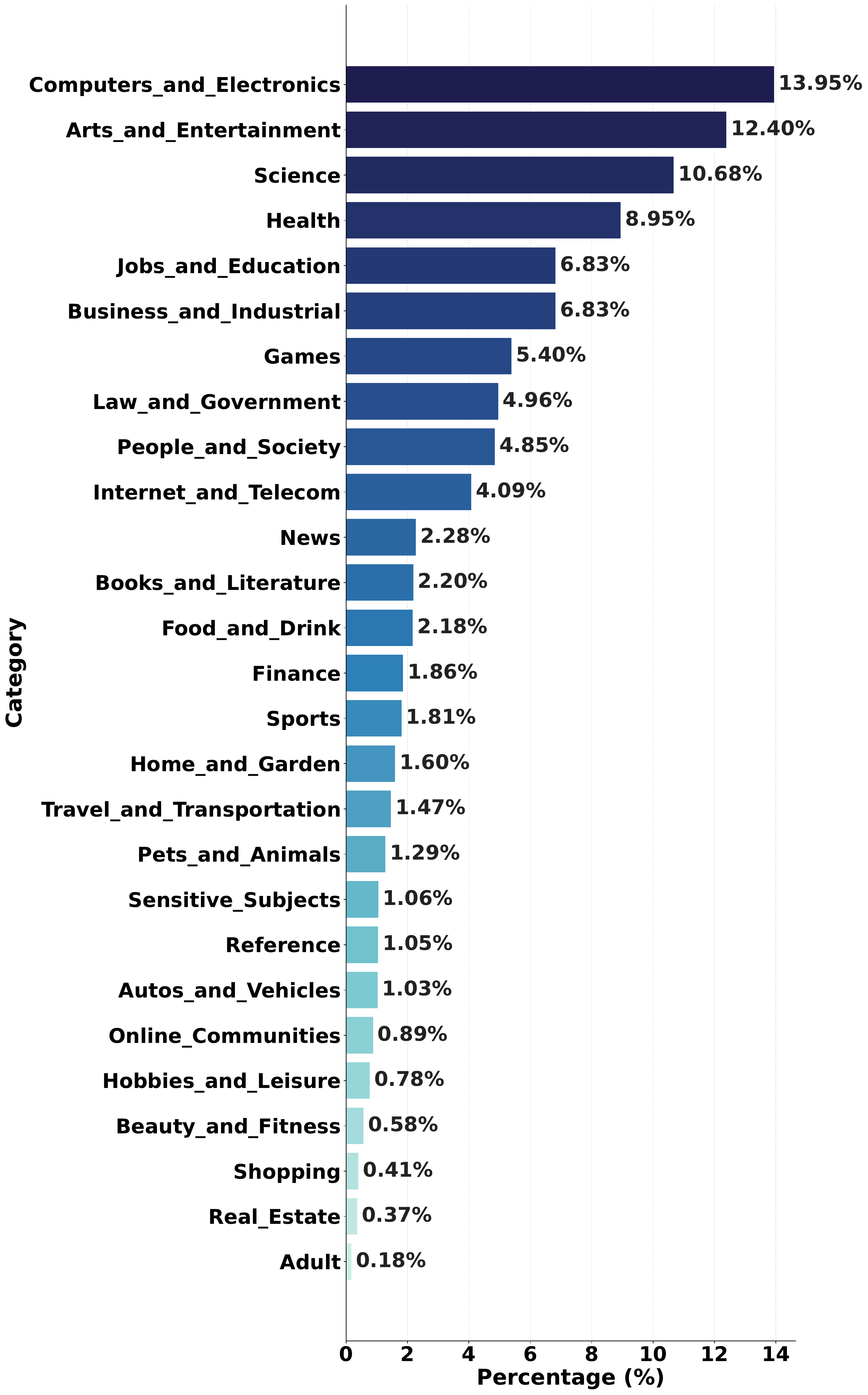}\\[-0.2em]
      {\small Dolma3}
    \end{minipage}
    
    \vspace{0.8em}
    
    \begin{minipage}[t]{0.31\textwidth}
      \centering
      \includegraphics[width=\linewidth]{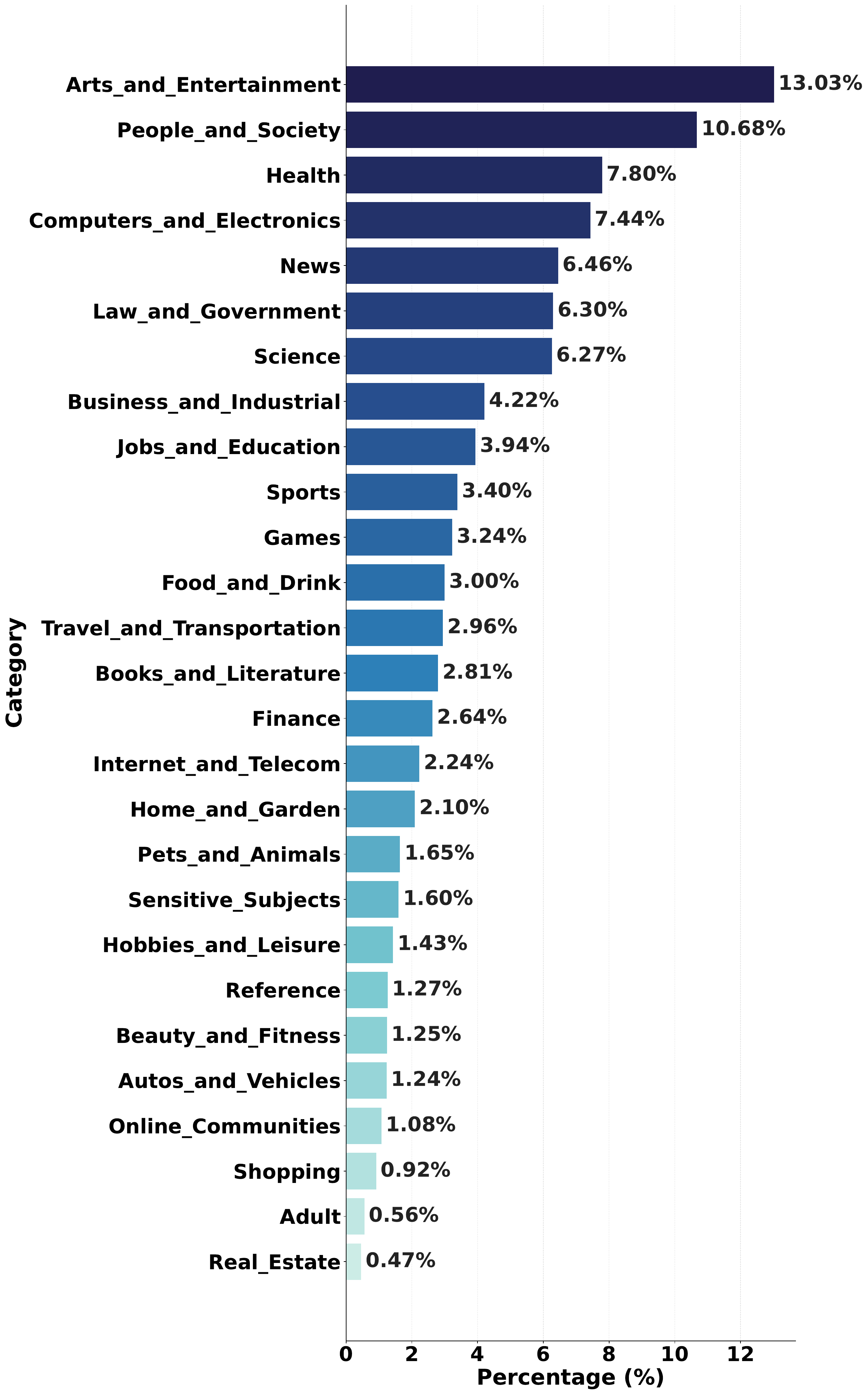}\\[-0.2em]
      {\small DCLM}
    \end{minipage}
    \hspace{0.08\textwidth}
    \begin{minipage}[t]{0.31\textwidth}
      \centering
      \includegraphics[width=\linewidth]{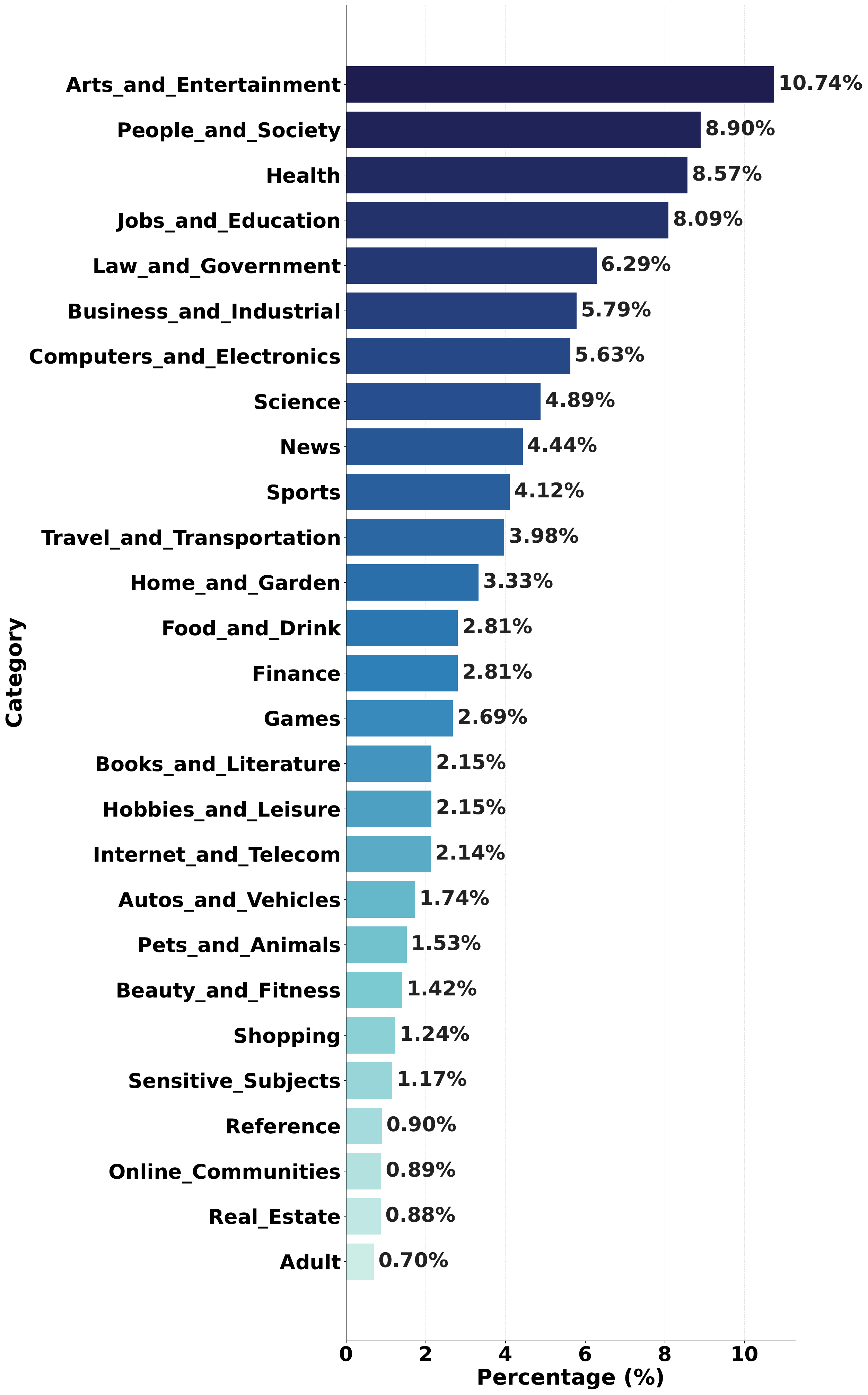}\\[-0.2em]
      {\small Nemotron-CC}
    \end{minipage}
  \end{center}
  
  \caption{Top-27 L1-domain distributions across five datasets under equal-volume (50M documents) comparison. CuraWeb exhibits the most balanced distribution across all domains, while FineWeb-Edu is heavily concentrated in Science (19.01\%) and Health (17.59\%).}
  \label{fig:cross_dataset_domains}
\end{figure}

\subsection*{F \quad Domain Distribution Evolution Across the Pipeline}
\begin{figure}[t] %
  \centering
  % ── 第一行：前三个阶段 ──────────────────────────────────
  \begin{subfigure}[b]{0.32\textwidth}
    \centering
    \includegraphics[width=\textwidth]{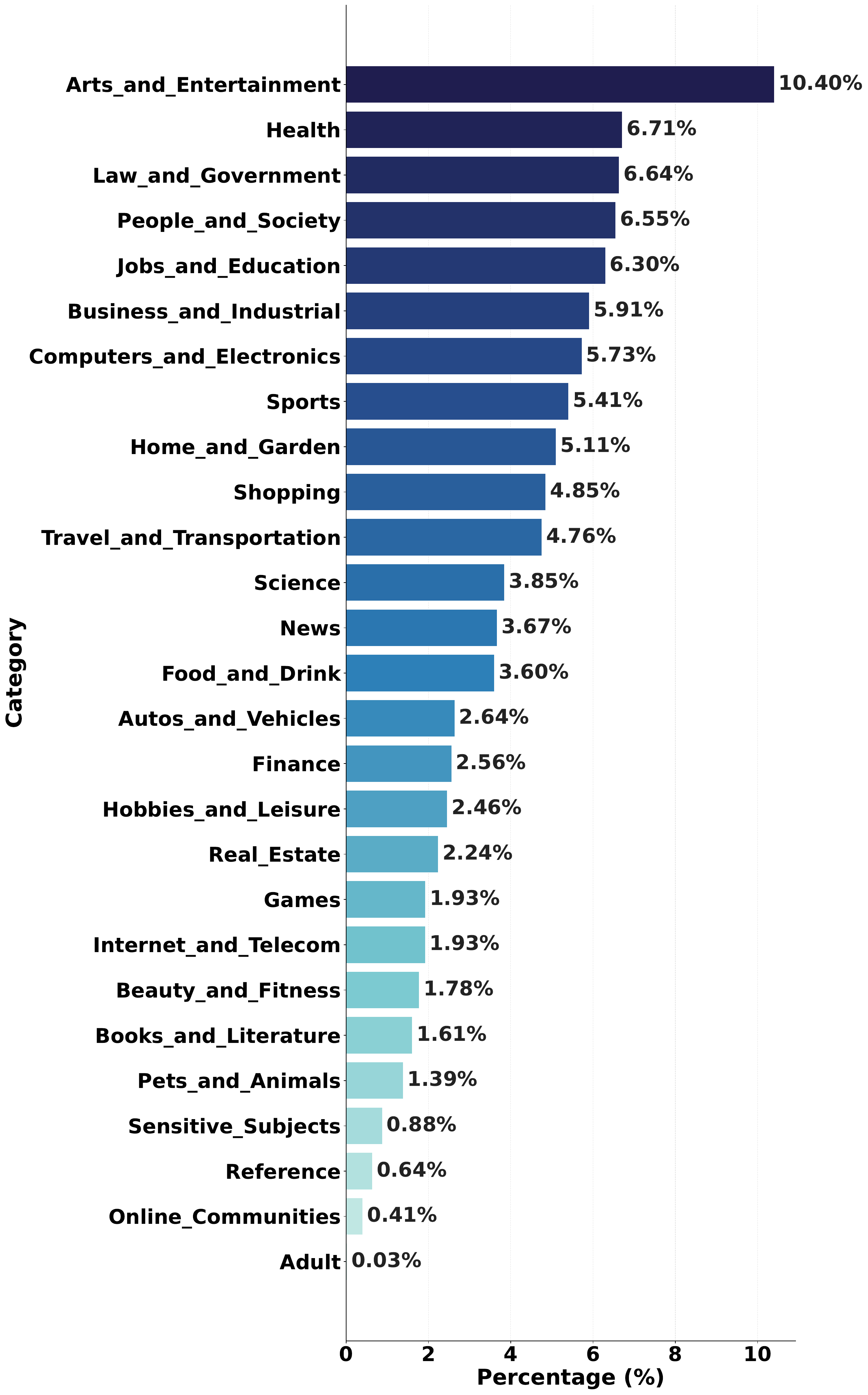}
    \caption{Rule Filtering}
    \label{fig:domain_rule}
  \end{subfigure}
  \hfill
  \begin{subfigure}[b]{0.32\textwidth}
    \centering
    \includegraphics[width=\textwidth]{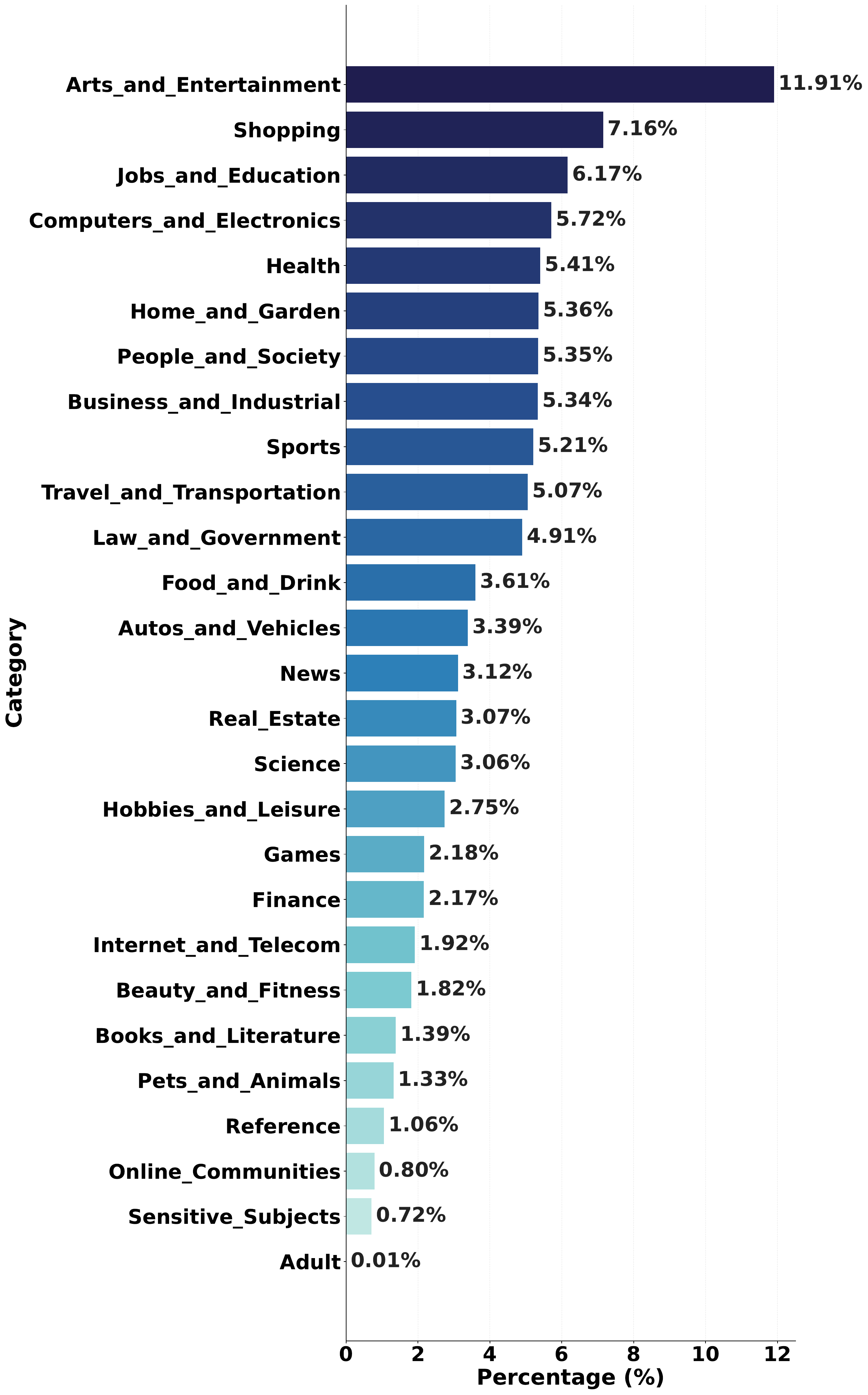}
    \caption{Model Filtering}
    \label{fig:domain_model}
  \end{subfigure}
  \hfill
  \begin{subfigure}[b]{0.32\textwidth}
    \centering
    \includegraphics[width=\textwidth]{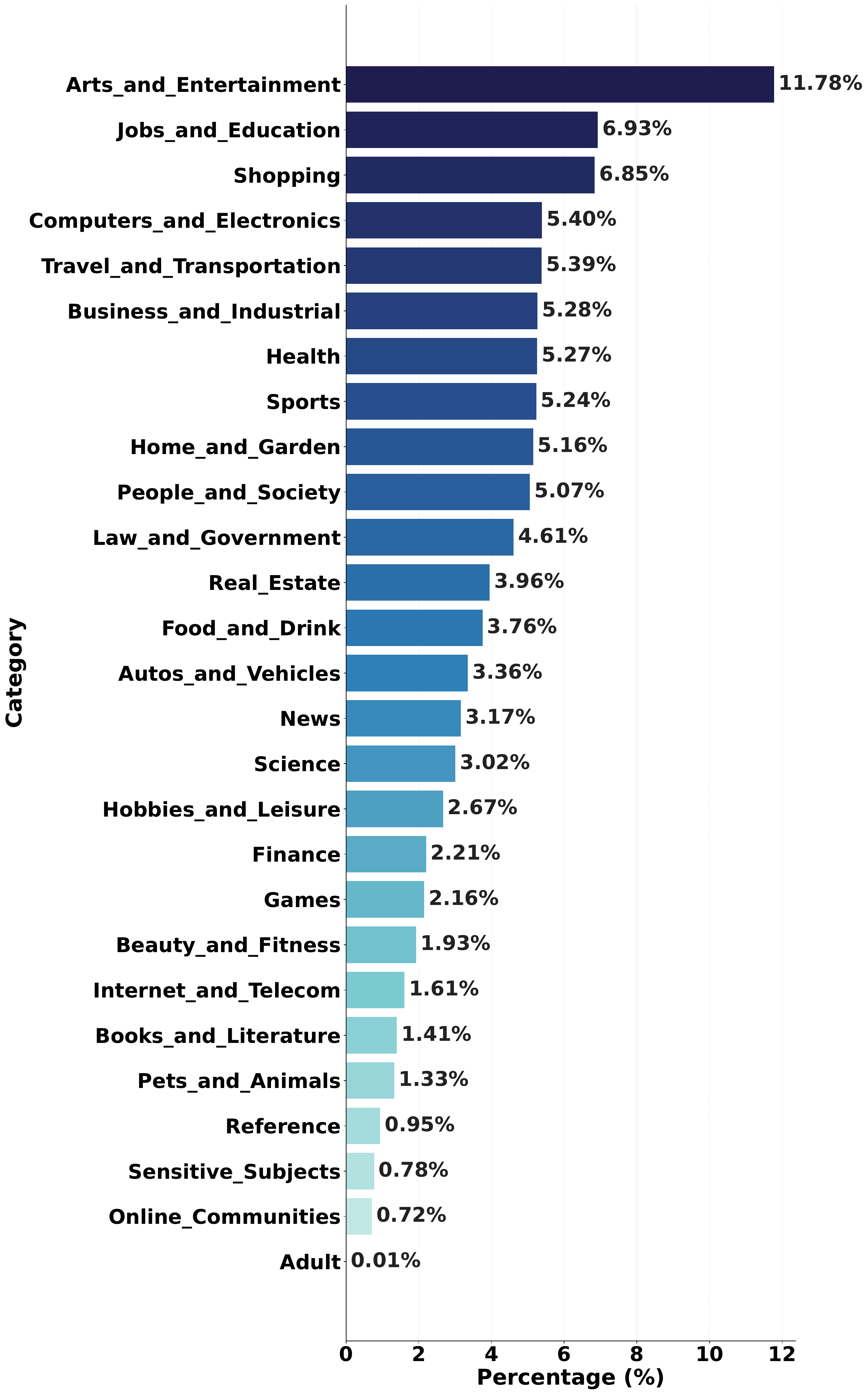}
    \caption{Fuzzy Dedup}
    \label{fig:domain_fuzzy}
  \end{subfigure}

  \vspace{0.8em} % 增加两行之间的垂直间距，避免视觉拥挤

  % ── 第二行：后三个阶段 ──────────────────────────────────
  \begin{subfigure}[b]{0.32\textwidth}
    \centering
    \includegraphics[width=\textwidth]{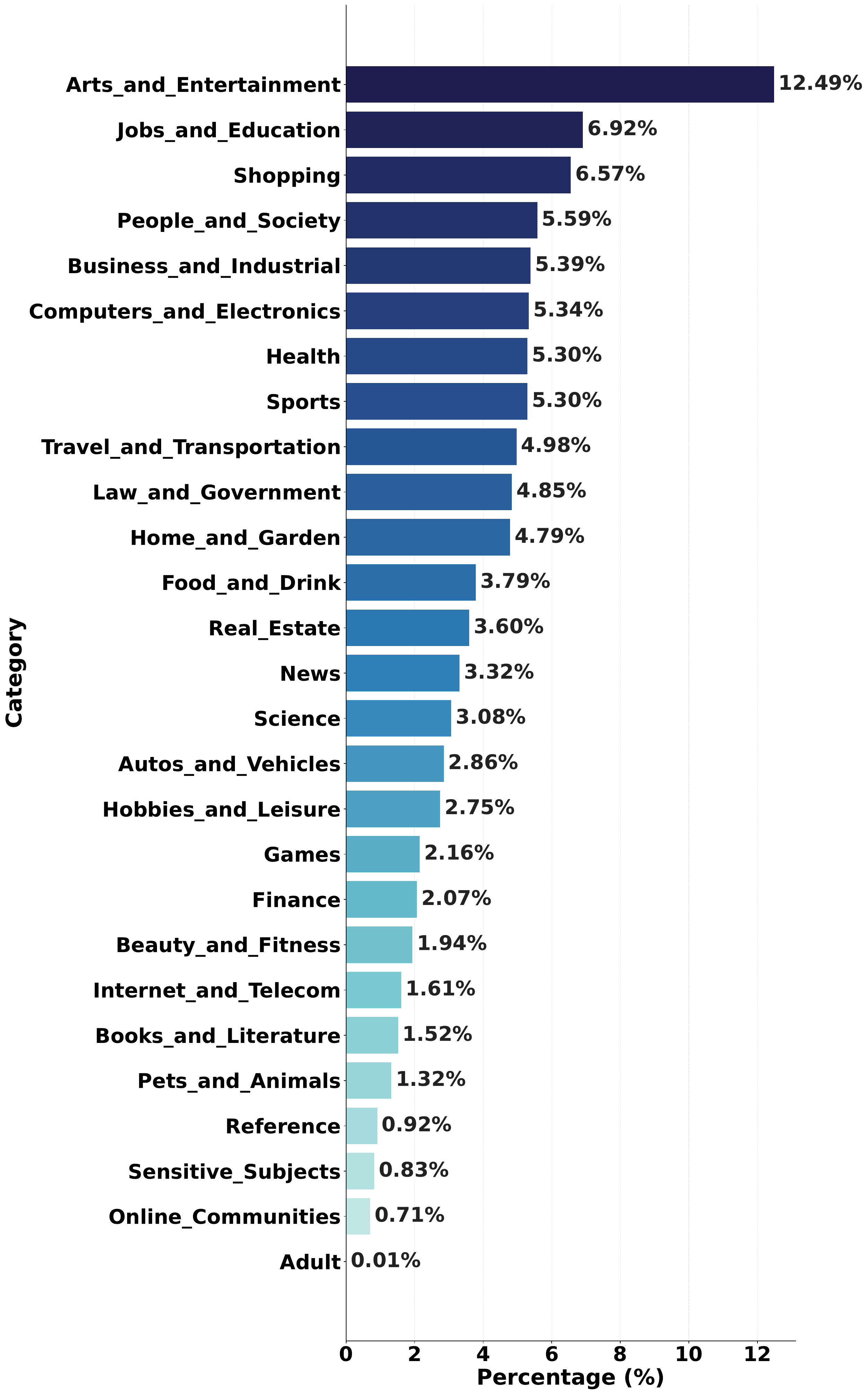}
    \caption{Semantic Dedup}
    \label{fig:domain_semantic}
  \end{subfigure}
  \hfill
  \begin{subfigure}[b]{0.32\textwidth}
    \centering
    \includegraphics[width=\textwidth]{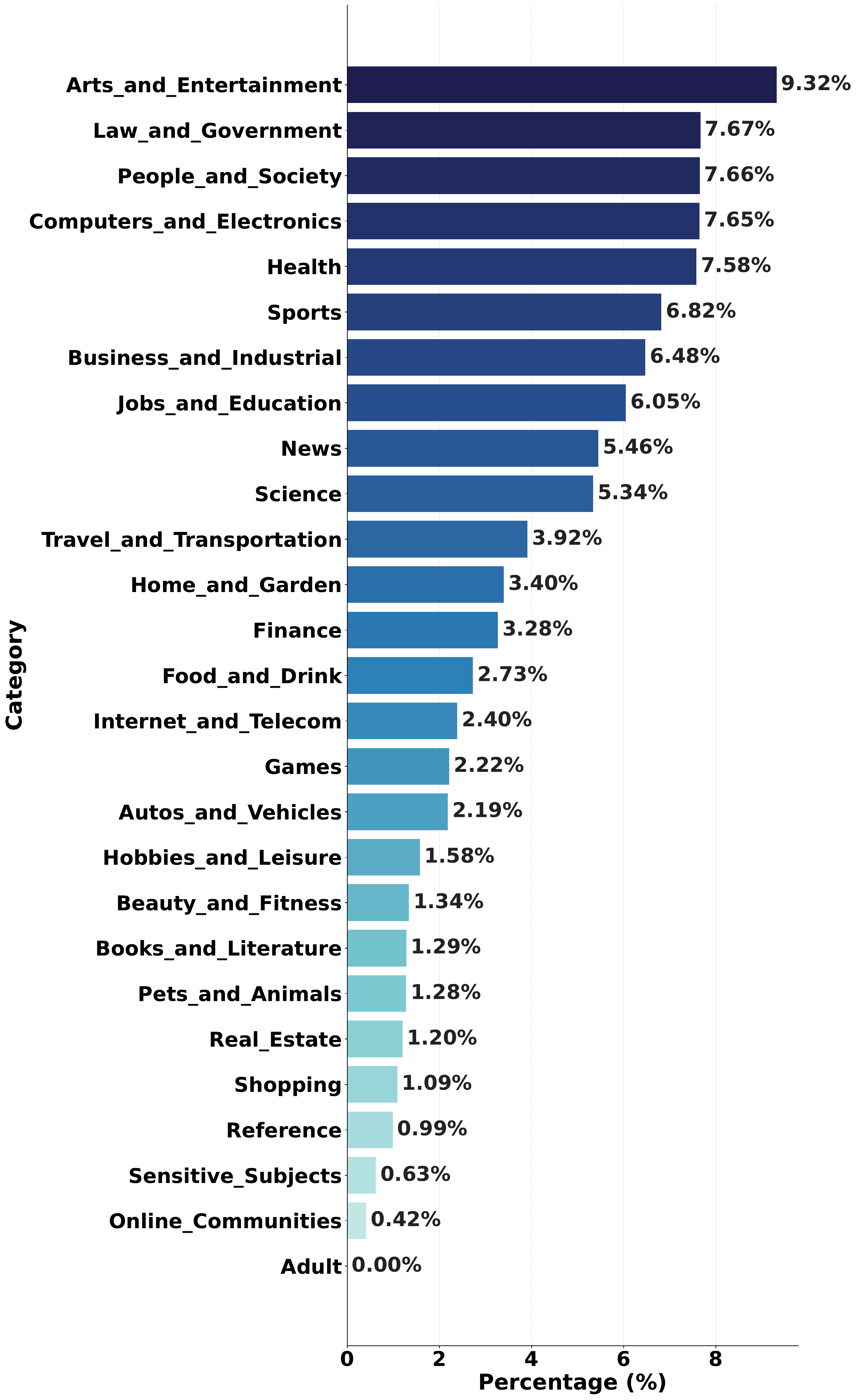}
    \caption{Value Filtering}
    \label{fig:domain_value}
  \end{subfigure}
  \hfill
  \begin{subfigure}[b]{0.32\textwidth}
    \centering
    \includegraphics[width=\textwidth]{Figures/lcc_pipeline_13-24_upgrade_pxsv2_sample5kw_config_6_20260512_163122_L1_top27.pdf}
    \caption{Importance Sampling}
    \label{fig:domain_sampling}
  \end{subfigure}

  \caption{Evolution of first-level domain distribution across the pipeline.}
  \label{fig:pipeline_domain_evolution}
\end{figure}

Figure~\ref{fig:pipeline_domain_evolution} illustrates the evolution of the first-level (L1) domain distribution across the CuraWeb pipeline. Throughout the progressive filtering and deduplication stages (Figures~\ref{fig:domain_rule}--\ref{fig:domain_value}), the distribution remains highly stable, with the Shannon entropy constrained tightly between $4.42$ and $4.43$. This ensures that our quality thresholds and semantic deduplication do not inadvertently suppress specific domains. In contrast, the final Importance Sampling stage (Figure~\ref{fig:domain_sampling}) successfully drives a controlled distribution shift, significantly enriching the corpus with high-value, knowledge-intensive domains such as $\texttt{Health}$, $\texttt{Science}$, and $\texttt{Jobs\_and\_Education}$ to enhance downstream model performance.

\end{document}